\documentclass{article}
\usepackage{arxiv} 
\usepackage[utf8]{inputenc} 
\usepackage[T1]{fontenc} 
\usepackage{hyperref} 
\usepackage{url} 
\usepackage{booktabs} 
\usepackage{amsfonts} 
\usepackage{nicefrac} 
\usepackage{
    microtype
} 
\usepackage{graphicx} 
\usepackage{amsmath, amsthm, amssymb, amsbsy, bm, mathtools, textcomp}
\usepackage[rflt]{floatflt} 
\usepackage{hyperref} 
\usepackage{xcolor} 
\usepackage{subcaption} 
\usepackage{adjustbox} 
\usepackage{tabularx} 
\usepackage{array} 
\usepackage{siunitx} 
\usepackage{comment} 
\usepackage[numbers]{natbib} 
\usepackage[textsize=tiny]{todonotes} 
\usepackage[listings]{
    tcolorbox
} 
\usepackage{url} 
\usepackage{breqn} 
\usepackage{wrapfig} 
\usepackage[final]{pdfpages} 
\usepackage{soul} 
\usepackage{enumitem} 
\usepackage{multirow} 
\usepackage{algorithm}
\usepackage{algpseudocode} 

\title{VAE-DNN: Energy-Efficient Trainable-by-Parts Surrogate Model For Parametric Partial Differential Equations}

\author{ Yifei Zong \\ 
School of Water Conservancy and Transportation \
Zhengzhou University \\ 
Zhengzhou, China, 450001 \\
Department of Civil and Environmental Engineering \ University of Illinois Urbana
Champaign \\ Urbana, IL 61801 \\ 
\texttt{yifeizong@outlook.com} \\ \And Alexandre Tartakovsky
\\ Department of Civil and Environmental Engineering \ University of Illinois Urbana Champaign \\
Urbana, IL 61801 \\ \texttt{amt1998@illinois.edu} \\ }

\begin{document}
\maketitle
\begin{abstract}

We propose a trainable-by-parts surrogate model for solving forward and inverse parameterized nonlinear partial differential equations. Like several other surrogate and operator learning models, the proposed approach employs an encoder to reduce the high-dimensional input $y(\bm{x})$ to a lower-dimensional latent space, $\bm\mu_{\bm\phi_y}$. Then, a fully connected neural network is used to map $\bm\mu_{\bm\phi_y}$ to the latent space, $\bm\mu_{\bm\phi_h}$, of the PDE solution $h(\bm{x},t)$. Finally, a decoder is utilized to reconstruct $h(\bm{x},t)$. The innovative aspect of our model is its ability to train its three components independently. This approach leads to a substantial decrease in both the time and energy required for training when compared to leading operator learning models such as FNO and DeepONet. The separable training is achieved by training the encoder as part of the variational autoencoder (VAE) for  $y(\bm{x})$ and the decoder as part of the $h(\bm{x},t)$ VAE. We refer to this model as the VAE-DNN model. 
VAE-DNN is compared to the FNO and DeepONet models for obtaining forward and inverse solutions to the nonlinear diffusion equation governing groundwater flow in an unconfined aquifer. Our findings indicate that VAE-DNN not only demonstrates greater efficiency but also delivers superior accuracy in both forward and inverse solutions compared to the FNO and DeepONet models. 

\end{abstract}

\section{Introduction}\label{sec:intro}

Modeling natural systems, such as subsurface flow, requires solving partial differential equations (PDEs). 
The computational cost of solving PDEs increases exponentially with the domain size and dimensionality. In some applications, including model inversion and uncertainty quantification, numerous simulations must be conducted. 
Machine learning (ML) deep neural operators learning in functional spaces and surrogate models operating in finite-dimensional subspaces have been proposed to reduce the computational time of evaluating forward solutions. However, the time and energy required for training ML models for large-scale applications can be very large. Here, we propose a trainable-by-parts surrogate model and demonstrate that it has a lower computational cost and higher accuracy than the state-of-the-art operator learning models. 

Neural operators, such as the Fourier neural operator (FNO) \cite{li2020fourier,wen2022uFNO} and the deep operator network (DeepONet)\cite{liu2021deep}, learn mappings from the parameter to the state (solution) of the PDE, although both parameter and state functions are discretized in the training datasets. In contrast, ML surrogate models learn a relationship between discretized parameters and solutions. Since the dimensionality of the discretized solutions can be very high, most surrogate models rely on dimensionality reduction to identify latent spaces of parameters and state variables, and then construct a mapping between these latent spaces. A popular approach for dimension reduction is to use linear (in latent variables) models such as the  Karhunen-Loeve expansion (KLE) in the KL-DNN method \cite{wang2024bayesian} and principal component analysis (PCA) in the PCA-Net method \cite{bhattacharya2021model}. A direct benefit of dimensionality reduction is that it reduces the number of parameters in the mapping network, thereby mitigating overfitting when the training data size is small. 

PCA, KLE, and linear autoencoders \cite{bao2020regularized}, approximate a function as a linear combination of fixed basis functions learned from the training dataset (ensemble of possible realizations of the function). While effective for systems with an inherently linear relationship between the function and its latent space, they may not provide an effective reduction for functions that are nonlinearly related to their latent spaces. For example, consider samples $\{ g^{(i)}(x) \}_{i=1}^{N}$ generated with the Gaussian function
$$g(x)=\frac{1}{\sigma\sqrt{2\pi}}\exp\left[- \frac{(x-\mu)^2}{\sigma^2}\right]$$
by sampling $\sigma$ and $\mu$ from independent distributions. Knowing the generating function, these samples can be represented exactly with the two-dimensional latent space $(\mu,\sigma)$. However, a linear model like KLE would require an infinite number of latent variables to exactly represent the samples $g^{(i)}(x)$. Examples of natural systems that can be represented more compactly with nonlinear dimension-reduction models include the permeability of fractured porous media and other systems with discontinuous properties \cite{seidman2022nomad}. 

Here, we propose the VAE-DNN method, a nonlinear reduced-order surrogate modeling framework that combines variational autoencoders (VAEs) with fully connected DNNs. VAE~\cite{kingma2013auto} is a probabilistic autoencoder that learns nonlinear, low-dimensional latent representations of high-dimensional data. It employs an encoder to compress input data into a latent distribution (typically a Gaussian distribution) and a decoder to reconstruct data sampled from this latent distribution. Training the VAE involves minimizing the loss function, which combines the reconstruction error with a Kullback-Leibler divergence term. In the VAE-DNN framework, we use two separate VAEs: one for spatially varying PDE parameters and the other for the state variables. A DNN is used to map the mean of the latent space of the PDE parameters to the latent space of state variables. The three components of the VAE-DNN model can be trained separately, which offers several advantages. First, it eliminates the need for large computer memory to load the entire training dataset, which consists of samples of the PDE parameters and the corresponding PDE solutions; for large-scale problems, each sample has a large number of degrees of freedom. Second, each component can be trained more easily without overfitting than models that train the DNN encoder, decoder, and map components jointly. 
Finally, the modular nature of VAE-DNN facilitates transfer learning, allowing the transfer of some of its components for modeling the system under new conditions \cite{zong2025mathematicsdigitaltwinstransfer}.

We compare the accuracy and efficiency of VAE-DNN to those of FNO and DeepONet for the nonlinear diffusion equation describing flow in an unconfined aquifer  \cite{freyberg1988exercise}. We demonstrate that VAE-DNN requires less time and energy for training and produces more accurate forward and inverse solutions than FNO and DeepONet. Our paper is organized as follows. We formulate the VAE-DNN method in Section \ref{sec:methodology}, compare VAE-DNN to FNO and DeepONet for the nonlinear diffusion equation, and present results in Section \ref{sec:results}. Conclusions are given in Section \ref{sec:conclusion}. The software implementation employed for numerical experiments is publicly available at \url{https://github.com/geekyifei/VAE-DNN_surrogate}.

\section{Novelty of this work} 

\begin{itemize}

\item In \cite{laloy2017inversion}, a VAE has been employed to construct a low-dimensional representation of channelized aquifers, where the trained VAE encoder is used to facilitate Bayesian inversion. However,  the VAE encoding was not used to construct a surrogate model as is done in our work. 

\item Several existing models used an encoder-DNN-decoder architecture. For example, in \cite{zhu2018bayesian}, the encoder is employed to extract low-dimensional features from high-dimensional input parameter fields, which are then passed through a DNN and transformed to the solution using a decoder. In DeepOnet, the branch and trunk networks act as encoder and decoder, respectively. In these models, all components must be trained jointly. The VAE-DNN method enables the training of all components independently, thereby reducing the time and energy required for training.  

\item VAE-DNN generalizes KL-DNN and PCA-Net models that use a similar trainable-by-parts architecture but rely on linear autoencoders based on KLE and PCA, respectively. 

\end{itemize}

\section{Methodology}\label{sec:methodology}
In Section \ref{sec:vae_dnn_surrogate}, we present the  VAE-DNN surrogate model to approximate solutions of parametric time-dependent PDE systems as a function of the space-dependent PDE parameters. To facilitate the VAE-DNN model presentation, we consider the general PDE problem:
\begin{align}
    \label{eq:general_pde}\mathcal{L}(h(\mathbf{x},t), y(\mathbf{x})) = 0, \quad \mathbf{x} \in \Omega, \quad t\in T,
\end{align}
subject to the initial condition
\begin{align}
    h(\mathbf{x},t=0)=h_{0}(\mathbf{x}),  \quad \mathbf{x} \in \Omega
\end{align}
and the boundary conditions
\begin{align}
    \label{eq:BC}\mathcal{B}(h(\mathbf{x},t))=g(\mathbf{x},t) \quad \mathbf{x}\in \Gamma, \quad t\in T,
\end{align}
where $\mathcal{L}$ is a known differential operator, $\Omega \subset \mathbb{R}^n \ (n = 1, 2, 3)$ is the spatial domain, $T$ is the time domain, $h(\mathbf{x},t)$ is the state (solution) of the PDE, $y(\mathbf{x})$ is the space-dependent parameter (property) of the PDE, $h_{0}(\mathbf{x})$ is the initial condition, $\mathcal{B}$ is the boundary condition operator, $g(\mathbf{x},t)$ is boundary function, and $\Gamma$ is the domain boundary. 

In Section \ref{sec:inverse_problem}, we formulate an inverse solution of this PDE problem using the VAE-DNN model.

\subsection{VAE-DNN as a Surrogate Model}\label{sec:vae_dnn_surrogate}

Consider the operator $\mathcal{G}^{\dagger}$ mapping  the parameter function $y(\mathbf{x})$ to the PDE solution $h(\mathbf{x}, t)$, i.e. $h = \mathcal{G}^{\dagger}(y)$. The operator $\mathcal{G}^{\dagger}$ can be approximated with a parametric surrogate model $\mathcal{G}^*$ trained using a labeled dataset consisting of $N_{\text{train}}$ pairs of the different realizations of  $y(\mathbf{x})$ and the corresponding $h(\mathbf{x}, t)$ solutions. The $y(\mathbf{x})$ and $h(\mathbf{x}, t)$ functions are usually nodal values on a spatial/spatio-temporal mesh. For example, in the numerical example considered in Section \ref{sec:results},  $y(\mathbf{x})$ is approximated on a two-dimensional $N_{x_1}\times N_{x_2}$ uniform mesh and is given by a two-dimensional tensor $\mathbf{y} \in \mathbb{R}^{N_{x_1} \times N_{x_2}}$. The numerical solution $h$ is obtained on the same mesh at $N_t$ time intervals and is given in the training dataset by a three-dimensional tensor $\mathbf{h} \in \mathbb{R}^{N_t \times N_{x_1} \times N_{x_2}}$. The resulting training dataset takes the form $D = \left\{  \mathbf{y}^{(i)} \rightarrow \mathbf{h}^{(i)}  \right\}_{i=1}^{N_{\text{train}}} $.

For complex PDE problems, the problem dimensionality (i.e., $N_{x_1}\times N_{x_2}$ and $N_t$) can be very large, requiring an exponentially larger number of surrogate model parameters. In turn, the required number of samples $N_{\text{train}}$ in the training dataset increases with the number of surrogate model parameters, which can make the construction of surrogate models for large-dimensional PDE problems impractical. 

\begin{figure}
    \centering
    \includegraphics[width=0.99\linewidth]{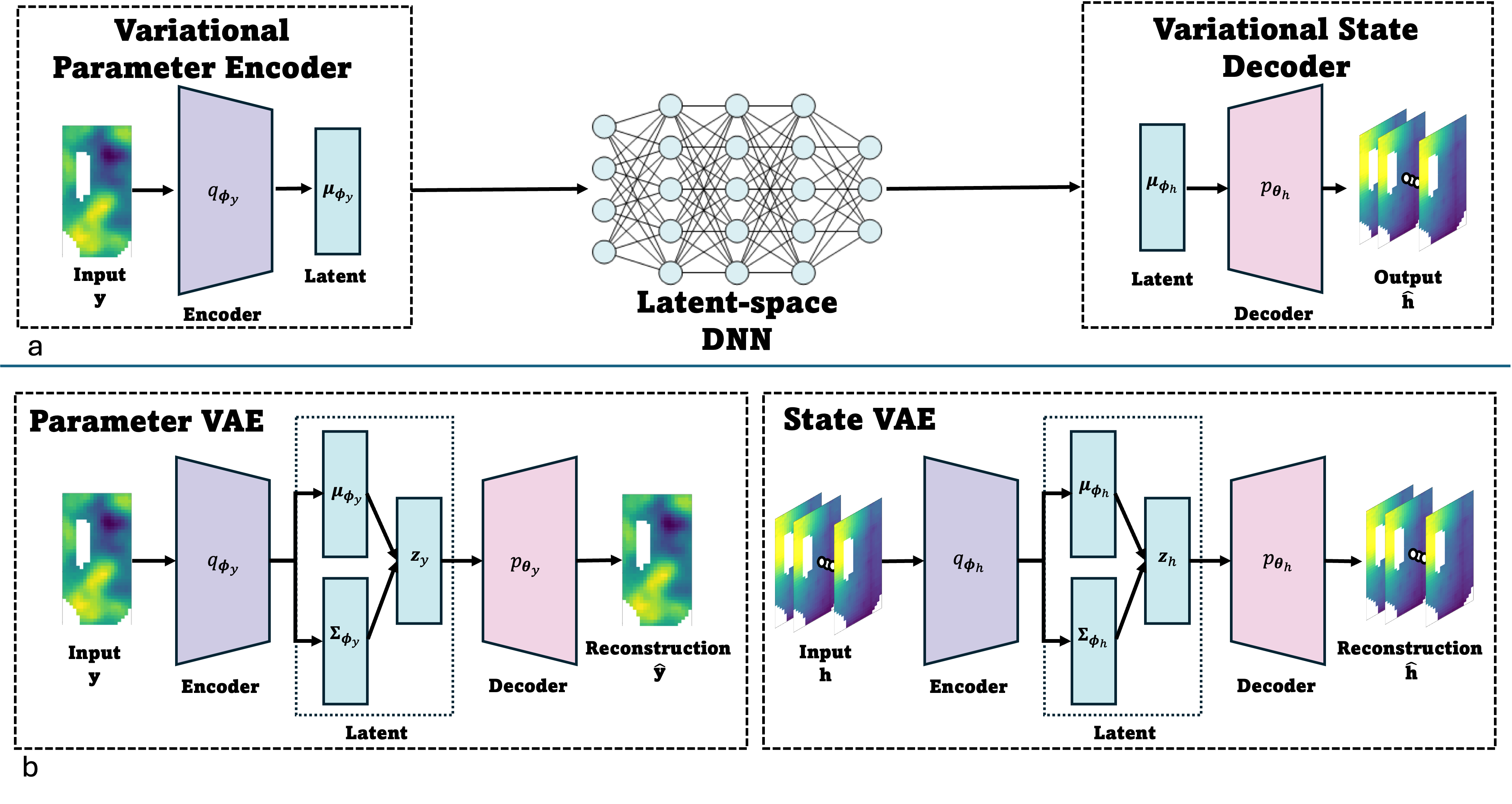}
    \caption{(a) Schematics of the VAE-DNN surrogate model: The $y$-VAE encoder, $q_{\phi_y}$, reduces the spatially varying $y$ field to its latent vector $\boldsymbol\mu_{\phi_y}$, a fully connected feed-forward DNN maps $\boldsymbol\mu_{\phi_y}$ to the latent vector $\boldsymbol\mu_{\phi_h}$ of the state variable, which is transformed to the state variable using the $h$-VAE decoder, $p_{\phi_h}$. (b) The encoder $q_{\phi_y}$ and the decoder $p_{\phi_h}$ are trained separately as parts of the $y$ and $h$ VAEs.}
    \label{fig:vae_dnn_schematics}
\end{figure}

Dimensionality reduction is a standard approach for reducing the computational cost of constructing surrogate models. Here, we propose to project $\mathbf{y}$ and $\mathbf{h}$  onto lower-dimensional latent vector spaces 
$\boldsymbol\mu_{\phi_y}$ and $\boldsymbol\mu_{\phi_h}$, 
respectively, using the encoder components of the $y$ and $h$ VAEs. Then, a DNN with a relatively small number of parameters is used to map $\boldsymbol\mu_{\phi_y}$ to $\boldsymbol\mu_{\phi_h}$:
\begin{equation}
    \boldsymbol\mu_{\phi_h}  \approx \mathcal{NN}(
   \boldsymbol\mu_{\phi_y}; \boldsymbol\psi),
\end{equation}
where $\boldsymbol\psi$ is the parameters of the fully-connected DNN. The VAE-DNN model is depicted in Figure \ref{fig:vae_dnn_schematics}.

 A VAE is an ML model consisting of the probabilistic encoder and decoder parts that are trained jointly. For the $h$ and $y$ variables in Eq \eqref{eq:general_pde}, the VAEs are trained using the $ \left\{ \mathbf{h}^{(i)} \right\}_{i=1}^{N_{\text{train}}}$ and $\left\{ \mathbf{y}^{(i)} \right\}_{i=1}^{N_{\text{train}}} $ data, respectively.  The VAE models are shown in Figure \ref{fig:vae_dnn_schematics}. Once trained, the $y$ encoder,  $q_{\phi_y}$, takes any sample $\bm{y}^*$ as an input and outputs its latent coordinates $\boldsymbol\mu^*_{\phi_y}$.  Then, the $y$ decoder, noted as $p_{\theta_y}$, takes $\boldsymbol\mu^*_{\phi_y}$ as an input and outputs $\hat{\bm{y}}^*$, which is a reconstruction of the field $\bm{y}^*$. The $h$ VAE performs similar operations with any sample of $h$. 
 
We note that VAE is a generative model and can be used to statistically reconstruct any vector. For example, in the $y$-VAE, in addition to  $\boldsymbol\mu_{\phi_y}^*$ (the mean of  latent distribution of $\bm{y}^*$ ), the encoder $q_{\phi_y}$ also outputs $\boldsymbol{\Sigma}_{\phi_y}^*$ (the covariance of the latent distribution of $\bm{y}^*$ ). The auxiliary random variable, 
\begin{align}
\mathbf{z} = \boldsymbol{\mu}^*_{\phi_y} + \boldsymbol{\Sigma}^{*1/2}_{\phi_y} \odot \boldsymbol{\epsilon},
\label{eq:latent_vector}
\end{align}
is then used as an input to the decoder, where $\boldsymbol{\epsilon}$ is a random sample from the normal distribution $\mathcal{N}(\mathbf{0}, \mathbf{I})$ and $\odot$ denotes element-wise multiplication. The output of the decoder will be a reconstructed $\hat{\bm{y}}^*$ corresponding to a sample of the latent variable from the normal distribution with the mean  $\boldsymbol\mu_{\phi_y}^*$ and covariance $\boldsymbol{\Sigma}_{\phi_y}^*$. The statistical sampling makes VAE a generative model. In this work, statistical sampling is only performed during training of the $h$ and $y$ VAEs and is not used in the inference of the VAE-DNN surrogate model. 

Once the $y$ and $h$ VAEs are trained. We use them to generate a latent space dataset $ \left \{ \boldsymbol{\mu}_{\phi_y}^{(i)} \rightarrow \boldsymbol{\mu}_{\phi_h}^{(i)} \right\}_{i=1}^{N_{\text{train}}}$ from the full dataset $D$. This step is achieved by using the pretrained VAE encoders as  $\boldsymbol{\mu}_{\phi_y}^{(i)}  = q_{\phi_y}( \mathbf{y}^{(i)} )$ and $\boldsymbol{\mu}_{\phi_h}^{(i)}  = q_{\phi_h}( \mathbf{h}^{(i)} )$. 
Then, the DNN can be trained by minimizing the mismatch between the latent mean vector $\hat{\boldsymbol\mu}_{\phi_h} = \mathcal{NN}(
   \boldsymbol{\mu}_{\phi_y}; \boldsymbol\psi)$ and the latent mean vector $\boldsymbol\mu_{\phi_h}$:
\begin{equation}\label{eq:loss_vaednn}
\boldsymbol{\psi}^{*}= \arg\min_{\boldsymbol\psi}\mathcal{L}
    (\boldsymbol{\psi}) = \frac{1}{N_{\text{train}}} \sum_{i=1}^{N_{\text{train}}}|| \mathcal{NN}(
   \boldsymbol{\mu}^{(i)}_{\phi_y}; \boldsymbol\psi)  - \boldsymbol{\mu}_{\phi_h}^{(i)} ||^{2}_{2}.
\end{equation}

In the inference step, VAE-DNN takes  $\boldsymbol{y}^*$ as an input, uses the $y$-VAE encoder to compute $\boldsymbol\mu^*_{\phi_y}$,  then the DNN to evaluate  $\boldsymbol\mu^*_{\phi_h}$, and the $h$-VAE decoder to estimate $\hat{\bm{h}}^*$. 

In Section \ref{sec:results}, we compare the VAE-DNN surrogate model with the state-of-the-art neural-operator models, the DeepONet and the FNO, for forward and inverse nonlinear diffusion PDE problems. The details of the FNO and DeepONet models are summarized in Appendix \ref{sec:appendix_fno} and \ref{sec:appendix_deeponet}, respectively.

VAEs provide nonlinear relationships between functions and the latent vectors, as opposed to linear autoencoders based on PCA and KLEs. Linear autoencoders have been used to construct surrogate models (e.g., the PCA-Net \cite{bhattacharya2021model} and KL-DNN \cite{wang2024bayesian}). The limitations of the linear autoencoders are mentioned in Section \ref{sec:intro}.

Several comments about the VAE training. Consider the $h$ VAE as an example. Given the unlabeled dataset $\mathcal{H}= \{\mathbf{h}^{(i)}\}_{i=1}^{N_{\text{train}}}$ with $N_{\text{train}}$ samples, the $h$ VAE is trained by minimizing the negative evidence lower bound (ELBO):
\begin{equation}\label{eq:vae_loss_empirical}
    \mathcal{L}(\phi_h, \theta_h ; \mathcal{H}) = 
    \frac{1}{N_{\text{train}}}\sum_{i=1}^{N_{\text{train}}} \left[ \left\| \mathbf{h}^{(i)}
    -  p_{\theta_h} (\mathbf{z}^{(i)}_h) \right\|_{2}^{2}  + \beta \left( \left
    \| \boldsymbol{\mu}_{\phi_h}(\mathbf{h}^{(i)})\right\|_{2}^{2}- \log \text{det}(\boldsymbol\Sigma_{\phi_h}
    (\mathbf{h}^{(i)})) + \text{Tr}(\boldsymbol\Sigma_{\phi_h}(\mathbf{h}^{(i)})) \right) \right]
\end{equation}
where 
$\mathbf{z}^{(i)}_h = \boldsymbol{\mu}_{\phi_h}(\mathbf{h}^{(i)}) + \boldsymbol{\Sigma}_{\phi_h}(\mathbf{h}^{(i)})^{1/2} \odot \boldsymbol{\epsilon}^{(i)}$, the first term is the data mismatch term, the second term is the KL regularizer, and $\beta$ is the regularization coefficient. Here, $\boldsymbol{\epsilon}^{(i)}$ is the $i$-th random sample from the normal distribution $\mathcal{N}(\mathbf{0}, \mathbf{I})$.

Decoding in VAE often diffuses the input image and fails to reconstruct high frequencies \cite{durall2020watch}. In \cite{jiang2021focal}, a Focal Frequency Loss (FFL) based on a Fourier frequency distance metric was introduced to improve the VAE accuracy. 

For a two-dimensional tensor $\mathbf{u}$ of $u(x_1, x_2)$ values computed on a uniform mesh $N_{x_1} \times N_{x_2}$, where $x_1$ and $x_2$ denote spatial coordinates, FFL is defined as follows.  The two-dimensional discrete Fourier transform $U(m, n)$ of $\mathbf{u}$ is:
\begin{align}
    U(m, n) = \sum_{k=0}^{N_{x_1}-1} \sum_{l=0}^{N_{x_2}-1} u(x^k_1, x^l_2) \cdot e^{-i 2\pi \left(\frac{mx^k_1}{N_{x_1}} + \frac{nx^l_2}{N_{x_2}} \right)},
\end{align}
where, $m$ and $n$ are the coordinates in the two-dimensional frequency domain, and $i$ is the imaginary unit. Let $U(m,n)$ and $\hat{U}(m,n)$ be the discrete Fourier transforms of the original data and reconstruction. The frequency error is computed as
\begin{align}
    E(m,n) = \left [ \left( \text{Re} \{ U(m,n) - \hat{U}(m,n) \} \right)^2 + \left( \text{Im} \{ U(m,n) - \hat{U}(m,n) \} \right)^2 \right ] ^{\frac{1}{2}}.
\end{align}
The weight matrix is defined as
\begin{align}
    W(m,n) = E(m,n)^{\alpha},
\end{align}
where $\alpha$ is a hyperparameter.
Then, FFL is computed as:
\begin{align}\label{eq:ffl_loss}
    \mathcal{L}_{\text{FFL}}(\phi, \theta ; \mathcal{U}) = \frac{1}{N_{\text{train}} N_m N_n} \sum^{N_{\text{train}}}_{i=1} \sum^{N_m-1}_{m=0} \sum^{N_n-1}_{n=0} W^{(i)}(m,n) \cdot | U^{(i)}(m,n) - \hat{U}^{(i)}(m,n)|^2,
\end{align}
where $N_m = N_{x_1}$, $N_n = N_{x_2}$ are the number of frequency components. The weight matrix $W(m,n)$ is computed dynamically during training based on the current frequency error, which forces the network to focus more on regions with higher frequency discrepancies. As a result, the final form of the loss function for VAE is a combination of Eq~\eqref{eq:vae_loss_empirical} and FFL loss~\eqref{eq:ffl_loss}:
\begin{align}\label{eq:loss_vae_final}
    \mathcal{L}_{\text{VAE}} = \mathcal{L} + \mathcal{L}_{\text{FFL}}.
\end{align}

For the considered examples, we found that adding $\mathcal{L}_{\text{FFL}}$ to $\mathcal{L}$ improves the accuracy of $h$-VAE but does not affect the accuracy of $y$-VAE. Therefore, in the following numerical examples, we use  $\mathcal{L}_{\text{VAE}}$ to train the $h$-VAE and $\mathcal{L}$ to train the $y$-VAE. 

\subsection{VAE-DNN inverse PDE solution}\label{sec:inverse_problem}

In this section, we propose a method for solving an inverse PDE problem using a pre-trained VAE-DNN surrogate model. In the inverse problem, we assume that $N_h^{\text{obs}}$ observations of $h$, $\{ \tilde{h}^i \}_{i=1}^{N_h^{\text{obs}}}$, and 
$N_{y}^{\text{obs}}$ measurements of $y$, $\left \{ \tilde{y}^i \right \}_{i=1}^{N_{y}^{\text{obs}}}$, are available. The coordinates of the $h$ and $y$ measurements are denoted as $\left \{ (\mathbf{x}^i_h, t^i_h) \right \}_{i=1}^{N_h^{\text{obs}}}$
and
$\left \{ \mathbf{x}^i_y \right \}_{i=1}^{N_{y}^{\text{obs}}}$, respectively. 

For simplicity, we assume that the observation locations remain constant over time. The solution of this inverse problem is $y^*(\mathbf{x})$ that minimizes a certain loss function. In this work, we define the inverse solution as maximum a posteriori (MAP), where the loss is defined as a function of  $\boldsymbol\mu_{\phi_y}$:

\begin{align}\label{eq:inverse_loss}
     \boldsymbol\mu_{\phi_y}^* = \arg\min_{ \boldsymbol\mu_{\phi_y}} \left \{ \frac{1}{N_h^{\text{obs}}} \sum_{i=1}^{N_h^{\text{obs}}} \left [ p_{\boldsymbol\theta_h}(\mathcal{NN}(  \boldsymbol\mu_{\phi_y}; \boldsymbol\psi))(\mathbf{x}^i_h, t^i_h) - \tilde{h}^i \right]^2 + \frac{1}{N_{y}^{\text{obs}}} \sum_{i=1}^{N_{y}^{\text{obs}}} \left [  p_{\boldsymbol\theta_y}(  \boldsymbol\mu_{\phi_y})(\mathbf{x}_y^i) - \tilde{y}^i \right]^2 + \gamma_{\text{inv}} \|  \mathcal{R}(  \boldsymbol\mu_{\phi_y} )  \|_2^2  \right \},
\end{align}
where $p_{\boldsymbol\theta_h}(\mathcal{NN}(  \boldsymbol\mu_{\phi_y}; \boldsymbol\psi))(\mathbf{x}^i_h, t^i_h)$ is the VAE-DNN predictions of $h$ at the sampling coordinate $ (\mathbf{x}^i_h, t^i_h)$, and $p_{\boldsymbol\theta_y}(  \boldsymbol\mu_{\phi_y})(\mathbf{x}_y^i)$ is $y(\mathbf{x}_y^i)$ reconstructed with  the $y$-VAE decoder. Here, $|| \mathcal{R}(  \boldsymbol\mu_{\phi_y}) ||_2^2$  denotes $\ell^2$ norm of the regularization function $\mathcal{R}(  \boldsymbol\mu_{\phi_y})$ and $\gamma_{\text{inv}}$ is the regularization coefficient. In this work, we employ Tikhonov regularization, where $\mathcal{R}(  \boldsymbol\mu_{\phi_y}) =    \boldsymbol\mu_{\phi_y}$ \cite{tikhonov1963solution}.
Regularization is required to make the solution of the otherwise ill-posed inverse problem unique. We note that $\boldsymbol\theta_h$, $\boldsymbol\theta_y$, and $\boldsymbol\psi$ are the pre-trained model parameters, which remain fixed when solving Eq \eqref{eq:inverse_loss}. In numerical examples, we assume that the $h$ and $y$ measurements are noise-free. Otherwise, the first and second terms in Eq \eqref{eq:inverse_loss} must be weighted by the coefficient inverse proportional to the variance of the observation noise \cite{ZONG2025randomized}.

After $  \boldsymbol\mu_{\phi_y}^*$ is estimated, $\mathbf{y}^*$ is reconstructed via the decoder of the $y$ VAE:
\begin{align}
    \mathbf{y}^* = p_{\boldsymbol\theta_y}(  \boldsymbol\mu_{\phi_y}^*).
\end{align}
Solving this inverse optimization problem with surrogate models, such as FNO, that directly maps $\mathbf{y}$ to $\mathbf{h}$ may lead to a less accurate solution because the dimensionality of $\mathbf{y}$ is significantly larger than that of $  \boldsymbol\mu_{\phi_y}$.  
The VAE-based dimensionality reduction provides a regularization in the inverse problem by reducing the number of unknown parameters, which constrains the inverse solution to a meaningful subspace.

\subsection{FNO and DeepONet inverse PDE solutions}
For comparison, we formulate MAP inverse solutions using the FNO and DeepONet neural operators. The FNO MAP estimate of $y$ can be found as the solution of the minimization problem: 
\begin{align}\label{eq:inverse_loss_fno}
    \mathbf{y}^* =  \arg\min_{\mathbf{y}} \Bigg\{  
      \frac{1}{N_h^{\text{obs}}} \sum_{i=1}^{N_h^{\text{obs}}} \left [ \mathcal{G}^*_{\text{FNO}}(\mathbf{y})(\mathbf{x}^i_h, t^i_h) - \tilde{h}^i \right]^2  
       + \frac{1}{N_{y}^{\text{obs}}} \sum_{i=1}^{N_{y}^{\text{obs}}} \left [ \mathbf{y}(\mathbf{x}^i_y) - \tilde{y}^i \right]^2    
       + \gamma_{\text{inv}} \|  \mathcal{R}(\mathbf{y})  \|_2^2 \Bigg\}
\end{align}
where $\mathcal{G}^*_{\text{FNO}}(\mathbf{y})(\mathbf{x}^i_h, t^i_h)$ is the prediction of the pretrained FNO model at $(\mathbf{x}^i_h, t^i_h)$ and  $\mathcal{R}(\mathbf{y}) = ||\mathbf{y}||_2^2$. We note that the dimensionality of this problem is higher than that of the minimization problem in the VAE-DNN inverse solution in Eq \eqref{eq:inverse_loss}.

The DeepONet model consists of the trunk and branch DNNs. The inputs to the trunk are the space and time coordinates of $h(\mathbf{x},t)$. The input to the branch DNN is the discretized parameter field $\mathbf{y}$ or its lower-dimensional representation. Following \cite{lu2021learning}, we approximate $y(\mathbf{x})$ with a truncated  Karhunen–Loève expansion (KLE):
\begin{align}
  \label{eq:parameter-KL}
  y(\mathbf{x}) = \overline{y}(\mathbf{x}) + \sum_{i=1}^{N_\xi} \sqrt{\lambda^{i}_y }\phi^i_y(\mathbf{x}) \xi^i
\end{align}
where $\overline{y}$ is the mean function, $\lambda^i_y$ and $ \phi_y^i$ are the eigenvalues and eigenfunctions of the $y$ covariance computed from training data, $N_\xi$ is the number of truncated KLE terms, and $\xi$ is the KLE mode. The details for KLE are described in Appendix \ref{sec:appendix_deeponet}. We use the truncated KLE modes $\boldsymbol\xi$ as input to the branch DNN. Then, the inverse problem is formulated as a minimization problem over $\boldsymbol{\xi}$:
\begin{align}\label{eq:inverse_loss_deeponet}
    \boldsymbol\xi^* =  \arg\min_{\boldsymbol\xi} \Bigg\{  
      \frac{1}{N_h^{\text{obs}}} \sum_{i=1}^{N_h^{\text{obs}}} \left [ \mathcal{G}^*_{\text{DON}}(\boldsymbol\xi)(\mathbf{x}^i_h, t^i_h) - \tilde{h}^i \right]^2  
      + \frac{1}{N_{y}^{\text{obs}}} \sum_{i=1}^{N_{y}^{\text{obs}}} \left [ y(\boldsymbol\xi)(\mathbf{x}^i_y) - \tilde{y}^i \right]^2    
       + \gamma_{\text{inv}} \|  \mathcal{R}(\boldsymbol\xi)  \|_2^2 \Bigg\}
\end{align}
where $\mathcal{G}^*_{\text{DON}}(\mathbf{y})(\mathbf{x}^i_h, t^i_h)$ is the prediction of the pretrained DeepONet model at $(\mathbf{x}^i_h, t^i_h)$, and $y(\boldsymbol\xi)(\mathbf{x}^i_y)$ is the KLE reconstruction of the parameter field evaluated at spatial locations $\mathbf{x}^i_y$. As in the VAE-DNN and FNO inverse solutions, we use Tikhonov regularization by setting $\|| \mathcal{R}(\boldsymbol\xi)  \|_2^2 = \||\boldsymbol\xi  \|_2^2 $.

\section{Application: parameter estimation for the Freyberg groundwater model}\label{sec:results}

\subsection{Problem description}\label{sec:problem_des}

We validate the proposed VAE-DNN framework on the Freyberg problem \cite{freyberg1988exercise}, a synthetic benchmark problem for parameter estimation in groundwater flow models. This problem considers an unconfined aquifer governed by the depth-averaged Darcy flow equation:
\begin{align}
    S_{y}\frac{\partial h(\boldsymbol{x},t)}{\partial t}= \nabla \cdot( K(\boldsymbol{x}) h(\boldsymbol{x},t) \nabla h(\boldsymbol{x},t)) +  f(t)+g(\boldsymbol{x},t) \label{eq:Darcy}
\end{align}
where $h(\boldsymbol{x},t)$ represents the hydraulic head $[L]$, $S_{y}$ is the specific yield $[-]$, $K(\boldsymbol{x})$ is the hydraulic conductivity $[LT^{-1}]$, $f(t)$ denotes the time-dependent recharge rate $[LT^{-1}]$ caused by precipitation, and $g(\boldsymbol{x},t)$ represents source/sink terms $[LT^{-1}]$ due to pumping wells and a river. 

\begin{figure}
    \centering
    \includegraphics[width=0.33\linewidth]{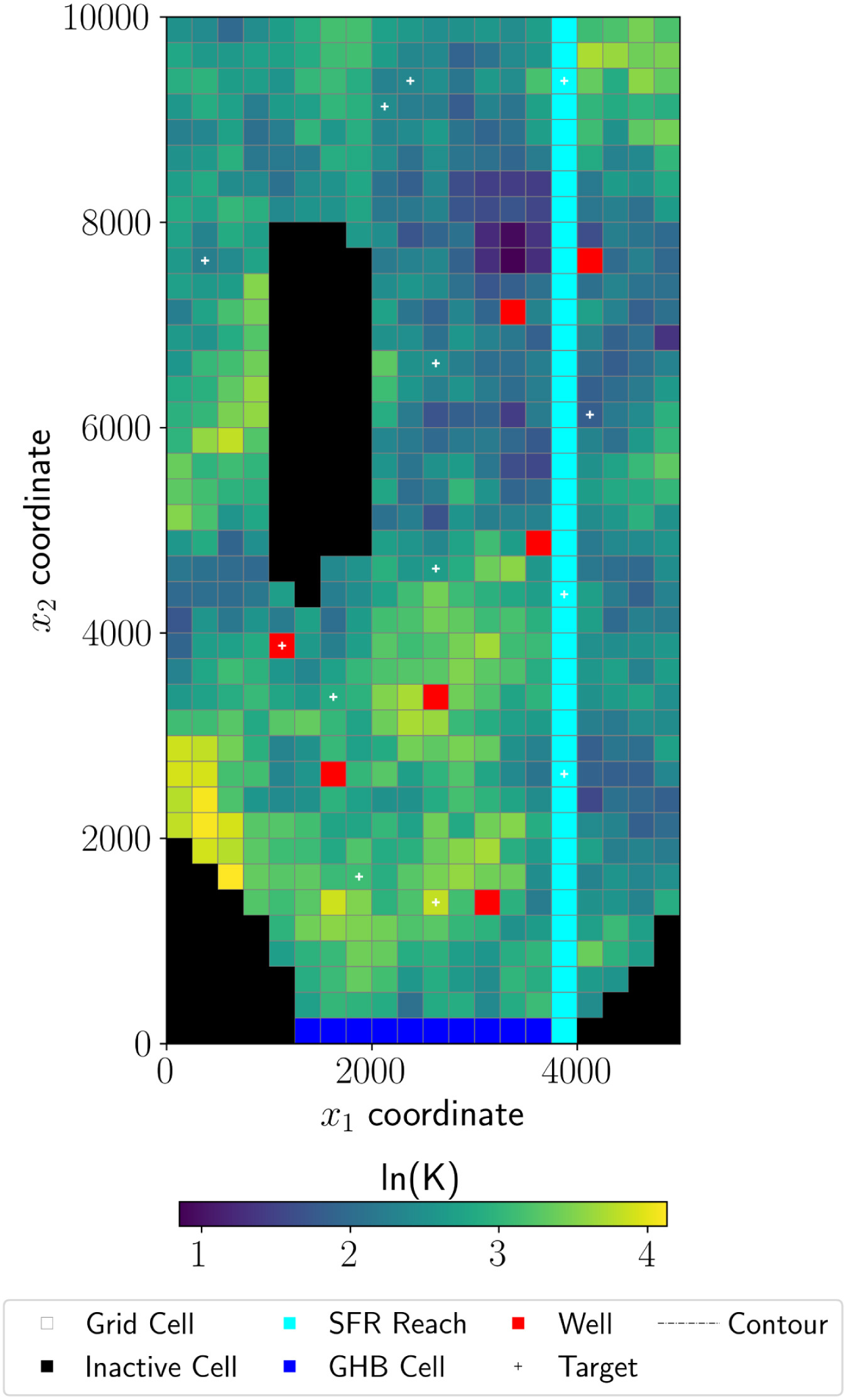}
    \caption{Computational model of the Freyberg problem implemented in MODFLOW 6. The reference log-hydraulic conductivity field is shown as the background colormap.}
    \label{fig:freyberg_conceptual}
\end{figure}

The model domain, shown in Figure \ref{fig:freyberg_conceptual}), spans $10000 \times 5000$ m. It is discretized with 40 $\times$ 20 uniform cells, each cell having dimensions of 250 m $\times$ 250 m.  There are $N = 706$ active cells. Black cells represent impermeable units forming a no-flow boundary. There are $N = 706$ active cells.  The domain is conceptualized as a headwater catchment with no-flow boundaries on the north, east, and west sides, while the southern boundary employs a Robin boundary condition (blue cells). A river (cyan cells) traverses the domain from north to south, and seven pumping wells (red cells) extract groundwater at specified rates. The initial condition is derived from the steady-state solution of Eq. \eqref{eq:Darcy} under the initial recharge rate $f_0 = f(t=0)$. We solve Eq. \eqref{eq:Darcy} using the MODFLOW 6 software. The Freyberg problem is run for 12 years using 25 time steps: an initial 10-year step followed by 24 monthly steps (1 month/step). Recharge rates vary monthly, and pumping rates vary annually during the final two years of operation.

To generate the $D$ dataset, we model the realizations of $K(\mathbf{x})$ as samples of a lognormal correlated random field $y(\mathbf{x})=\ln K(\mathbf{x})$ following the multi-scale procedure described in \cite{white2020toward}. Then, Eq \eqref{eq:Darcy} is solved for each sample of the $K$ field. The size of the training dataset is set to  $N_{\text{train}} = 4950$. 
A similar approach is used to generate the discretized reference fields $\mathbf{y}_{\text{ref}}$ and $\mathbf{h}_{\text{ref}}$, which are used to assess the accuracy of the forward and inverse solutions.

\subsection{Numerical Results}
In this section, we evaluate the performance of the proposed VAE-DNN surrogate model by comparing it with state-of-the-art neural operator-based surrogate models, namely DeepONet and FNO, on the Freyberg problem. We compare these models for both forward prediction and inverse modeling tasks. We use the relative $\ell^2$ error to quantify the accuracy of the predicted state field in the forward problem and the estimated parameter field in the inverse solution. The relative $\ell^2$ error between a predicted variable $\hat{\mathbf{u}}$ and the reference $\mathbf{u}$ is defined as  
\begin{eqnarray}\label{eq:l2_error}
    r\ell^2_u  &=& \left ( \frac{  || \hat{\mathbf{u}} - \mathbf{u} ||^2}{|| \mathbf{u} ||^2} \right )^{\frac{1}{2}}.
\end{eqnarray}

\subsubsection{Forward problem}
For the forward problem, the VAE-DNN, FNO, and DeepONet surrogate models are trained to predict $\mathbf{h}$ at $N_t$ time steps as a function of $\mathbf{y}=\ln\mathbf{K}$. The accuracy of forward prediction is evaluated by comparing with the reference state field $\mathbf{h}_{\text{ref}}$ generated by the MODFLOW simulation. As described in Section \ref{sec:problem_des}, the Freyberg dataset includes impermeable geological units that are known and fixed in their locations. In the training dataset, values of $y$ and $h$ corresponding to inactive cells are assigned a constant value of $-1$. To ensure that these inactive regions do not bias the learning process, a binary mask is applied during training to exclude inactive cells from contributing to the loss. This masking strategy is used for training all three surrogate models. Additionally, the input and output data are normalized using the mean and standard deviation across the training dataset as 
$$
\tilde{u}_i = \frac{u_i - \overline{u}_i}{\sigma_{u,i}}, 
$$
 where $u=y$ or $h$,  $u_i$ is the $i$th element of vector $\mathbf{u}$ (the value of $u$ at the $i$th cell), and  $\overline{u}_i$ and $\sigma_{u,i}$ are the sample mean and standard deviation of $u_i$, respectively. 

In the $y$ and $h$ VAEs, the latent space dimensions are set to $N^y_{\text{latent}} = 150$ and $N^h_{\text{latent}} = 90$, respectively. The DNN size is set to two hidden layers, each with 500 neurons. Two-dimensional convolutions are employed for the parameter field $\mathbf{y}$. For the spatio-temporal field $\mathbf{h}$, we compare two different approaches: the first approach utilizes three-dimensional convolutional layers, while the second applies two-dimensional convolutions to each temporal snapshot of $h$. A separate channel provides the time of each snapshot/convolution. The second approach is found to be slightly more accurate and approximately four times faster. Therefore, we present below only the results obtained with the second approach for constructing the $h$-VAE. The details of the VAE architecture design are discussed in Appendix \ref{sec:appendix_cvae}. The implementation details for FNO and DeepONet are given in Appendix \ref{sec:appendix_fno} and \ref{sec:appendix_deeponet}, respectively.

 In training all surrogate models, we set the batch size to 32 and use the Adam optimizer with a learning rate of \num{1e-4}. Table~\ref{tab:forward_results} summarizes the accuracy, training time, and energy consumption required for training the VAE-DNN, DeepONet, and FNO models. The accuracy is measured using the relative $\ell^2$ error, $r\ell^2_h$. To ensure a fair comparison, the total number of trainable parameters is kept similar across the three models. For the VAE-DNN model, we report the number of parameters, training time, and energy consumption for each of its three components as well as the total values for the entire model. 

\begin{table}[!htb]
    \centering
    \caption{Forward problem: $r\ell_h^2$ error in the $\textbf{h}$ estimates,  training times, and the energy consumption in VAE-DNN, DeepONet, and FNO models.}
    \begin{tabular}{cccccc}
         \toprule
         & $r\ell_h^2$ error & Number of & Total training & Total Energy & Total number\\
&  & parameters & time (sec) & consumption (Joules) &of epochs \\
 %
         \toprule
         VAE-DNN & \num{4.57e-04} &  \num{4848405} & \num{9.38e02}  & \num{9.10e04} & 1800 \\
         y-VAE &  &  \num{733805} & \num{1.88e02}  & \num{1.48e04} & 400 \\
          h-VAE &  &  \num{3 488 000} & \num{6.90e02}  & \num{7.10e04} & 1000 \\
           DNN &  &  \num{626 600} & \num{0.60e02}  & \num{0.52e04} & 400 \\
         \hline         
         DeepONet & \num{8.25e-04} & \num{4912080} & \num{3.06e05}  & \num{4.49e07} &  2000 \\
         \hline
         FNO & \num{4.59e-04} & \num{4767864} & \num{3.46e03} & \num{5.89e05} & 2000 \\
         \bottomrule
    \end{tabular}
    \label{tab:forward_results}
\end{table}

\begin{figure}[!htb]
    \centering
    \begin{subfigure}[b]{0.6\textwidth}
        \includegraphics[width=\textwidth]{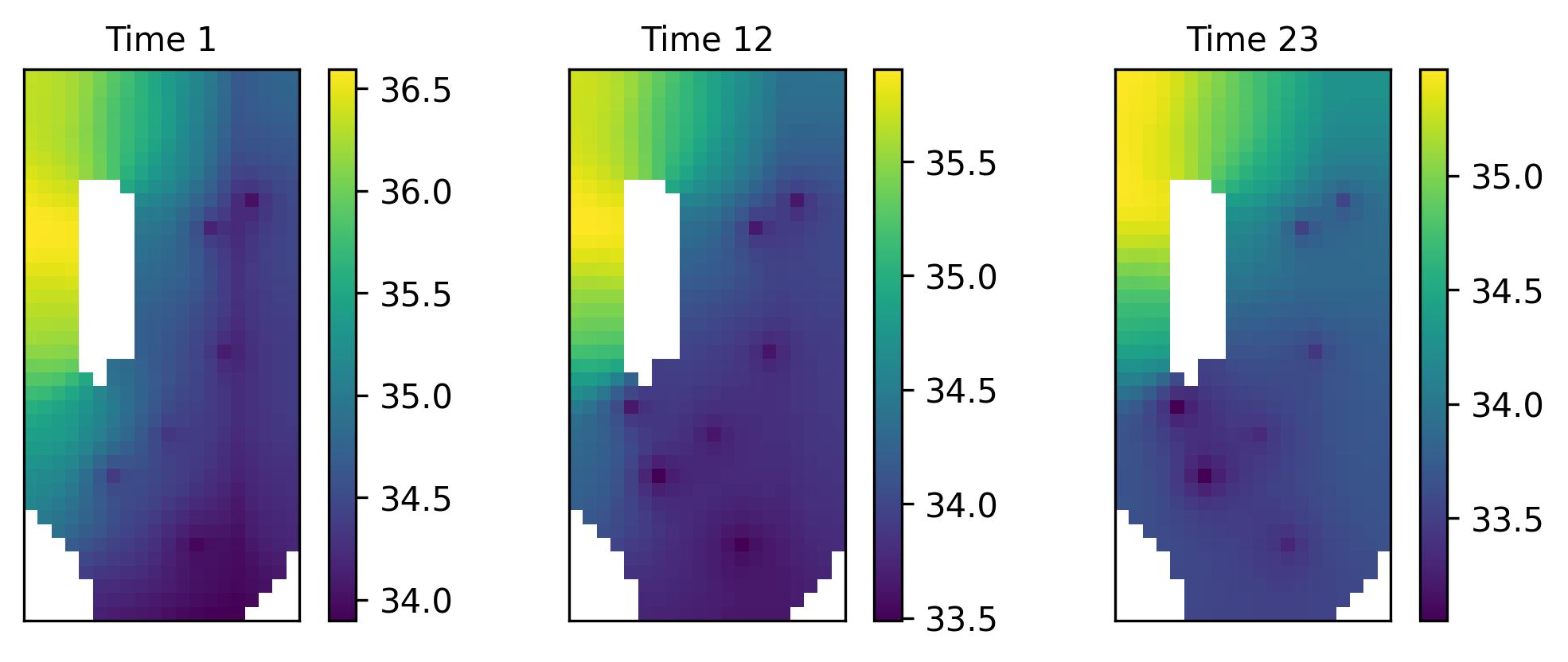}
        \caption{Reference}
    \end{subfigure}
    \begin{subfigure}[b]{0.6\textwidth}
        \includegraphics[width=\textwidth]{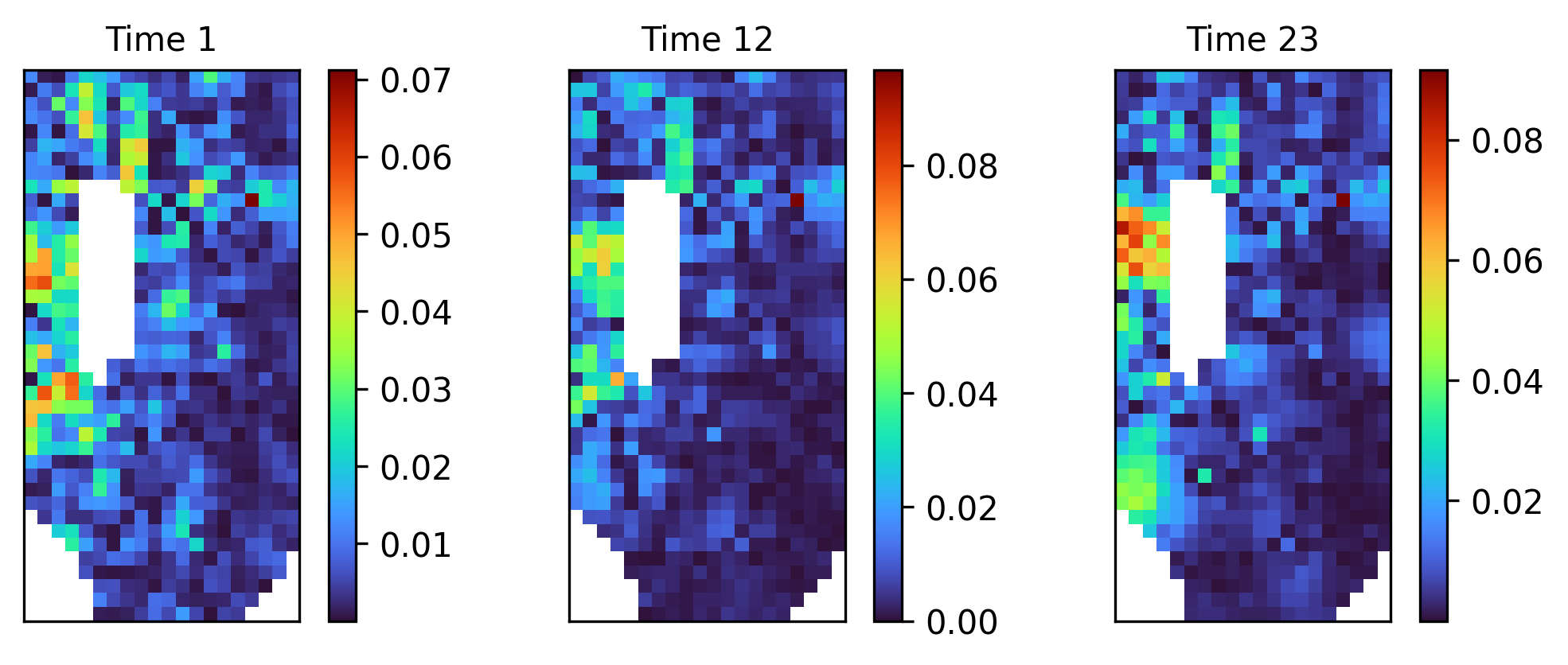}
        \caption{VAE-DNN}
    \end{subfigure}
    \begin{subfigure}[b]{0.6\textwidth}
        \includegraphics[width=\textwidth]{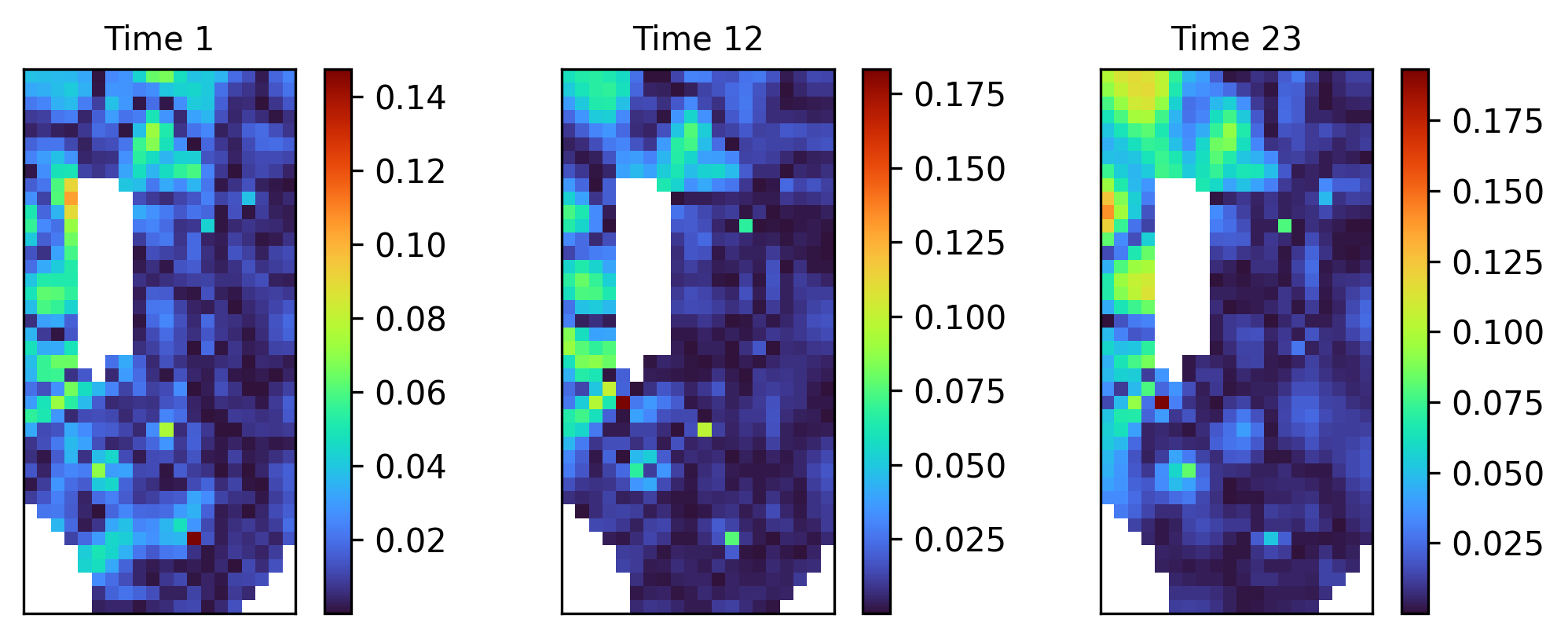}
        \caption{DeepONet}
    \end{subfigure}
        \begin{subfigure}[b]{0.6\textwidth}
        \includegraphics[width=\textwidth]{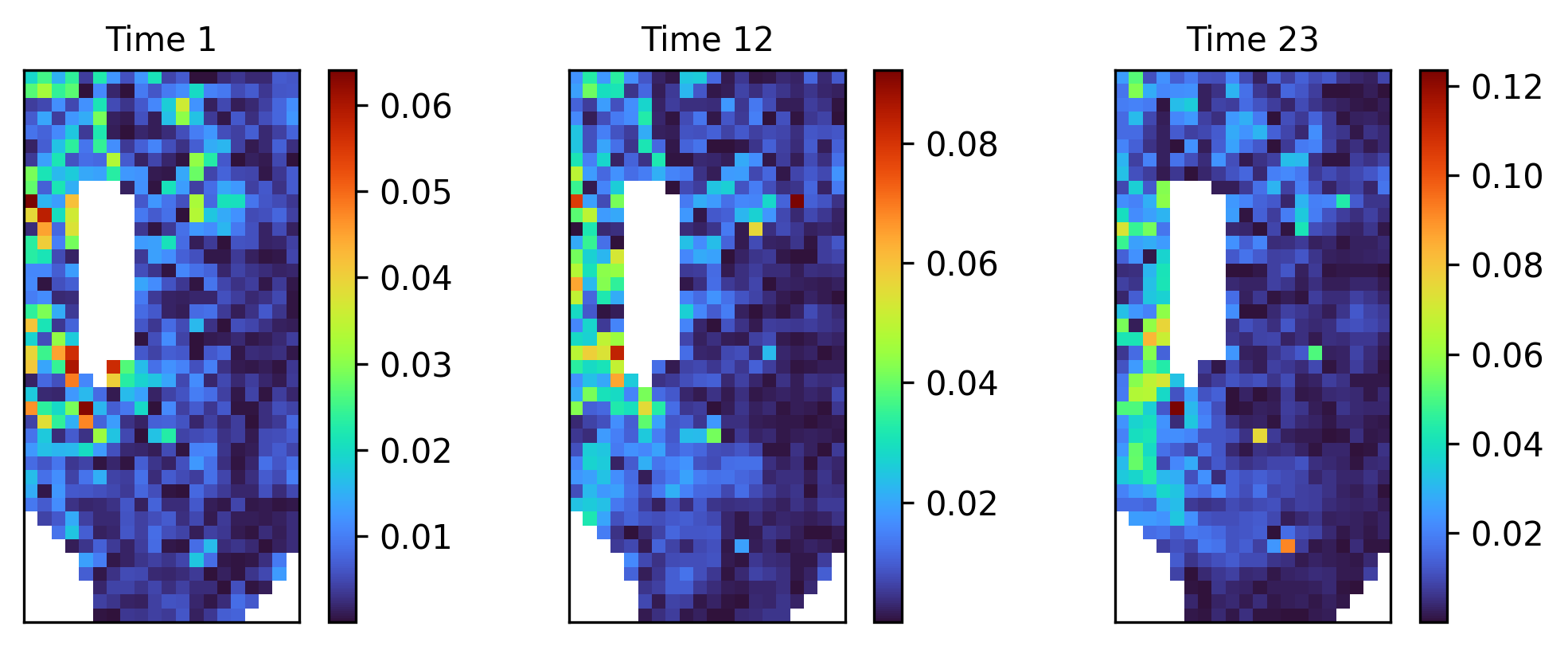}
        \caption{FNO}
    \end{subfigure}
    \caption{Forward Problem: (a) the reference hydraulic head field at three times (corresponding to timesteps 1, 12, and 23) and the point errors in the head predictions given by (b) VAE-DNN, (c) DeepONet, and (d) FNO. }
    \label{fig:forward_h}
\end{figure}

The $r\ell^2_h$ are of the same order in the three methods, with VAE-DNN achieving the lowest error $(4.57 \times 10^{-4})$, and DeepONet yielding the highest error $(8.25 \times 10^{-4})$. The training time and the energy consumption of VAE-DNN are the smallest among the three methods. Among the VAE-DNN components, the highest training time and energy consumption are incurred by h-VAE, and the lowest are by the DNN component. Figure~\ref{fig:forward_h} presents the reference state field $h_{\text{ref}}$ and the pointwise errors in $h$ predicted by the VAE-DNN, DeepONet, and FNO models at three time instances. DeepONet exhibits the largest maximum pointwise errors. The VAE-DNN incurs larger errors in regions where the hydraulic head is high at later times, whereas the FNO model displays higher errors near the well locations. Overall, these results highlight the effectiveness of the trainable-by-parts strategy and learning in a reduced latent space in the VAE-DNN, which together enable high accuracy while maintaining superior energy and computational efficiency.

\subsubsection{Inverse VAE-DNN, FNO, and DeepONet solutions}
Here, we use $N_y^{\text{obs}} = 13$ and  $N_h^{\text{obs}} = 13\times 25$ observations of $y(\mathbf{x})$ and $h(\mathbf{x})$ collected at 13 observation wells to estimate $\mathbf{y}$. The observations are drawn from $\textbf{y}_{\text{ref}}$ and $\textbf{h}_{\text{ref}}$.    

\begin{figure}
    \centering
    \includegraphics[width=0.5\linewidth]{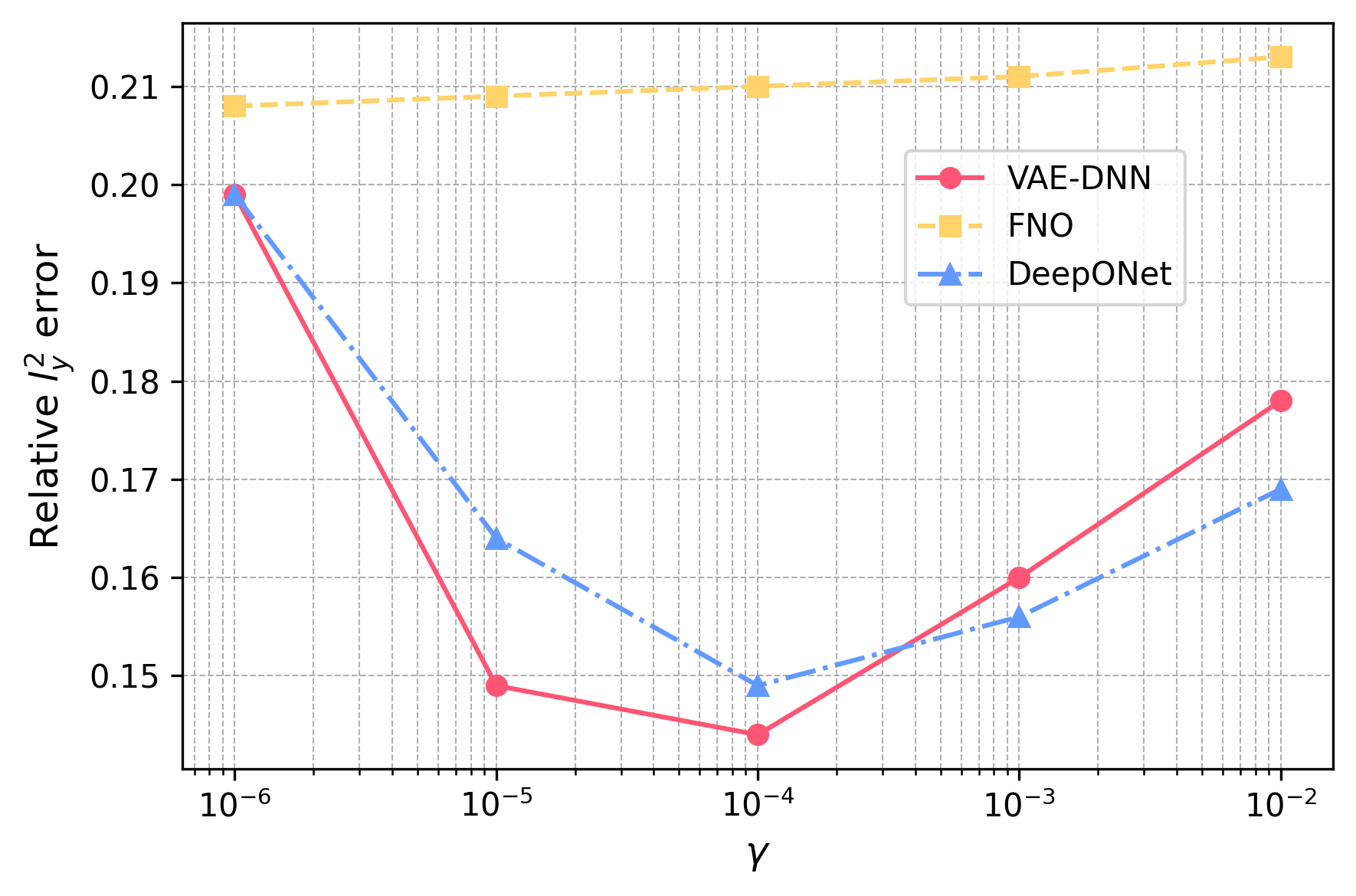}
    \caption{Relative $r\ell^2_y$ error in the estimated $y$ field as a function of the regularization parameter $\gamma$ for VAE-DNN, FNO, and DeepONet. }
    \label{fig:inverse_error_gamma}
\end{figure}

Figure~\ref{fig:inverse_error_gamma} presents the $r\ell^y_2$ errors in the inverse solutions as functions of the regularization parameter $\gamma_{\text{inv}}$. In DeepONet and VAE-DNN, $r\ell^y_2$ initially decreases with increasing $\gamma_{\text{inv}}$ and then increases. The optimal $\gamma_{\text{inv}}$ value in both methods is $10^{-4}$. The VAE-DNN achieves the lowest error of $\num{1.44e-01}$ versus $r\ell_2 = \num{1.49e-01}$ in DeepONet. 
The FNO error in the inverse solution is larger than in the other two methods ($r\ell_2 \approx \num{2.1e-01}$) and weakly depends on  $\gamma_{\text{inv}}$. 

\begin{figure}[!h]
    \centering
    \begin{subfigure}[b]{0.75\textwidth}
        \includegraphics[width=\textwidth]{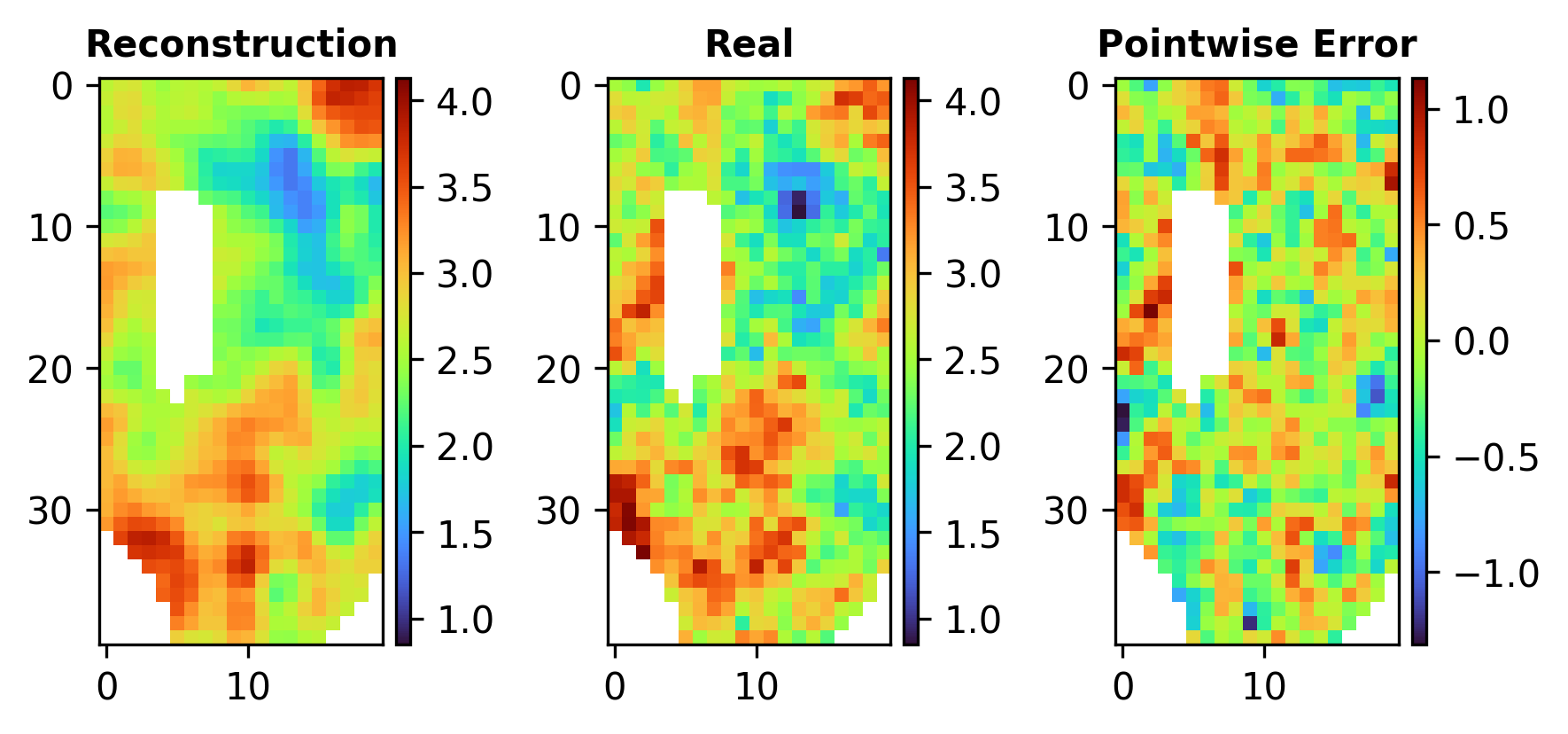}
        \caption{VAE-DNN}
    \end{subfigure}
    \begin{subfigure}[b]{0.75\textwidth}
        \includegraphics[width=\textwidth]{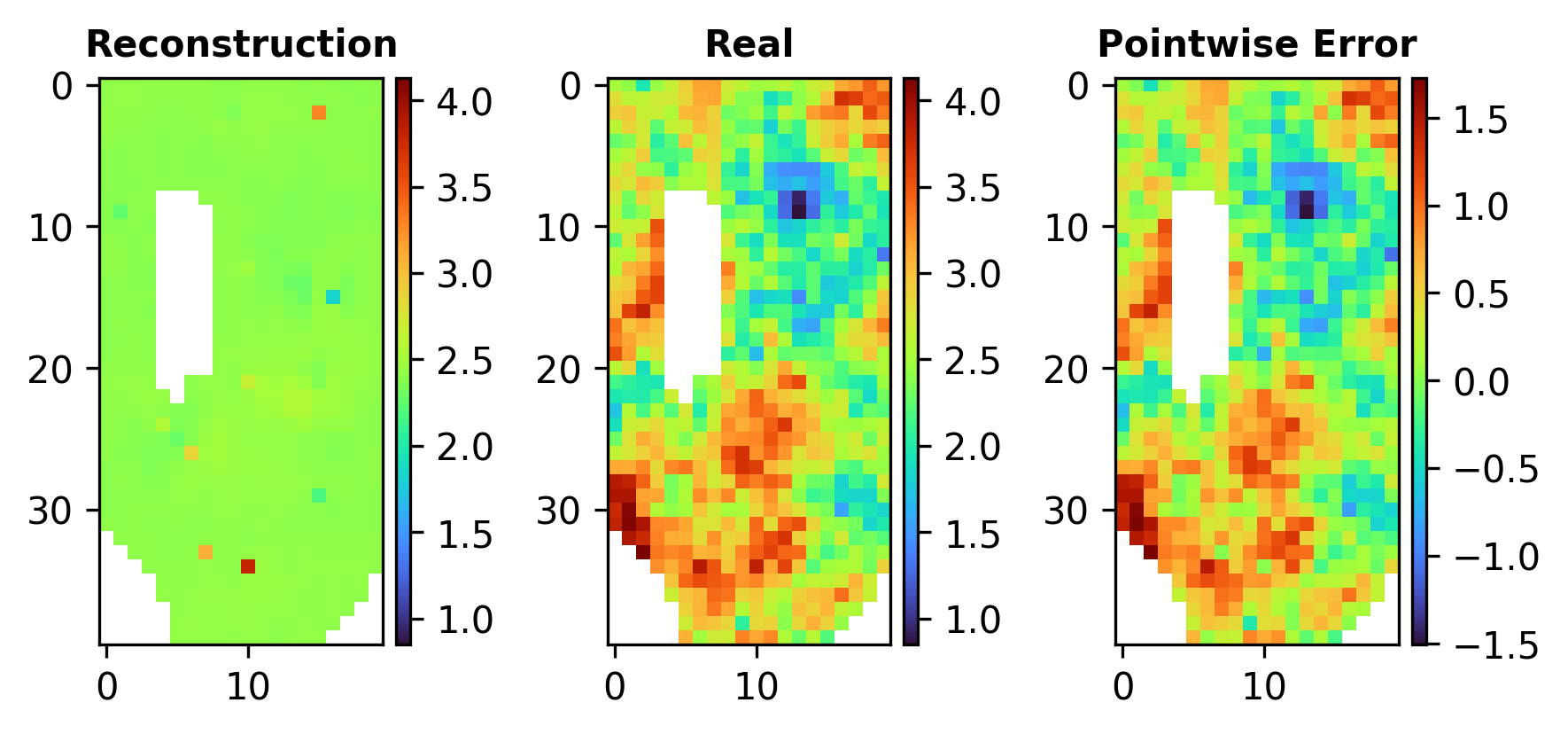}
        \caption{FNO}
    \end{subfigure}
    \begin{subfigure}[b]{0.75\textwidth}
        \includegraphics[width=\textwidth]{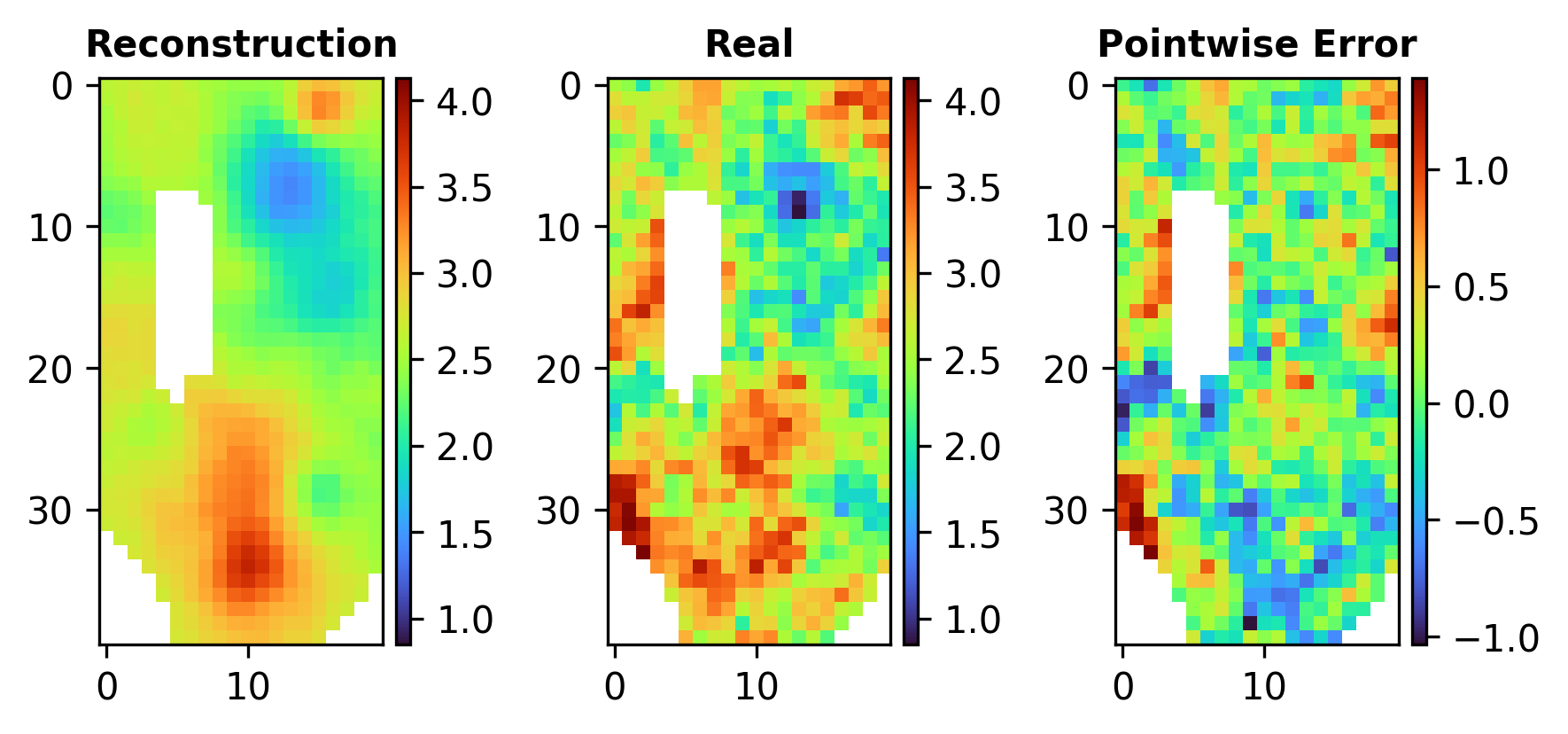}
        \caption{DeepONet}
    \end{subfigure}
    \caption{The estimated log hydraulic conductivity $\textbf{y}$, $\textbf{y}_{\text{ref}}$, and the point error $y_{\text{ref}}(\boldsymbol{x}) - {y}(\boldsymbol{x})$ in (a) VAE-DNN, (b) FNO, and (c) DeepONet.}
    \label{fig:inverse_y}
\end{figure}

Figure~\ref{fig:inverse_y} visualizes the estimated $\mathbf{y}$ in the three methods and the corresponding point errors. The VAE-DNN method achieves the smallest point errors, while FNO yields the largest point errors. The estimated $\mathbf{y}$ field in the FNO inverse solution is close to $\overline{\mathbf{y}}$ (the sample average in the training dataset), except for some deviations near the wells.
We note that the FNO surrogate model has a similar $r\ell^h_2$ error in the forward head solution to that of VAE-DNN and is approximately half the error of DeepONet. A relatively poor performance of the FNO in obtaining the inverse solution may be attributed to the fact that the FNO inverse is done in the full space of $y$, while the VAE-DNN and DeepONet find inverse solutions in the latent spaces of $y$. 
Consequently, the notably greater error observed in the FNO inverse solution can be attributed to the direct estimation of $y$ in the FNO model. In contrast, the other two models rely on inferring low-dimensional latent representations of $y$.

\section{Conclusions}\label{sec:conclusion}

In this work, we presented the trainable-by-parts VAE-DNN surrogate model for solving forward and inverse PDE problems. The model consists of three components: the parameter field encoder, the DNN, and the state variable decoder. Unlike DeepONet, CNN, and other models with a similar encoder-DNN-decoder design, the VAE-DNN model allows for training the three components separately. 
Our comparisons with the FNO and DeepONet models of the synthetic Freyberg aquifer system, governed by a nonlinear diffusion equation, demonstrated that this training-by-parts approach reduces training time and energy consumption, while improving the accuracy of both forward and inverse solutions. In addition, training by parts reduces the memory requirement because each component of the VAE-DNN model is trained on a subset of the training dataset. 

The parameter encoder and the state variable decoder are trained as parts of two VAEs. For a two-dimensional (space-dependent) parameter field, we use a standard convolutional VAE. For the three-dimensional (space-time) state variable, we employ a VAE architecture with two-dimensional convolutional layers, treating the temporal dimension as an additional channel. This design captures temporal dependencies while maintaining computational and energy efficiency. 

VAEs provide a nonlinear mapping from fields to their latent spaces, which has an advantage over linear auto-encoders based on PCA or KLE in finding more compact latent spaces for complex problems.  

Our results also show the importance of dimension reduction for solving inverse problems. In the VAE-DNN and DeepONet models, the inverse problem is solved in the latent space of the variables, while in FNO, the inverse solution is obtained in the full space. As a result, while the FNO model generates a forward solution that closely aligns with the VAE-DNN and outperforms DeepONet’s accuracy, its inverse solution is markedly less precise compared to the other two methods.

\section{Acknowledgements}
This research was partially supported by the Department of Energy (DOE)  Advanced Scientific Computing Research program and  CUSSP (Center for Understanding Subsurface Signals and Permeability), an Energy Earthshot Research Center funded by the DOE, Office of Science under FWP 81834. Pacific Northwest National Laboratory is operated by Battelle for the DOE under Contract DE-AC05-76RL01830. 

\bibliographystyle{unsrt}

\appendix

\section{Implementation details of the convolutional variational autoencoder}\label{sec:appendix_cvae}

In this section, we describe the $y$-VAE and $h$-VAE models for the Freyberg problem. We treat the discretized two-dimensional parameter field $\mathbf{y} \in \mathbb{R}^{ N_{x_1} \times N_{x_2}}$ as a two-dimensional tensor with one channel, and model it with a standard convolutional VAE using two-dimensional convolution layers. 

\begin{figure}[!h]
    \centering
    \begin{subfigure}[b]{0.45\textwidth}
        \includegraphics[width=\textwidth]{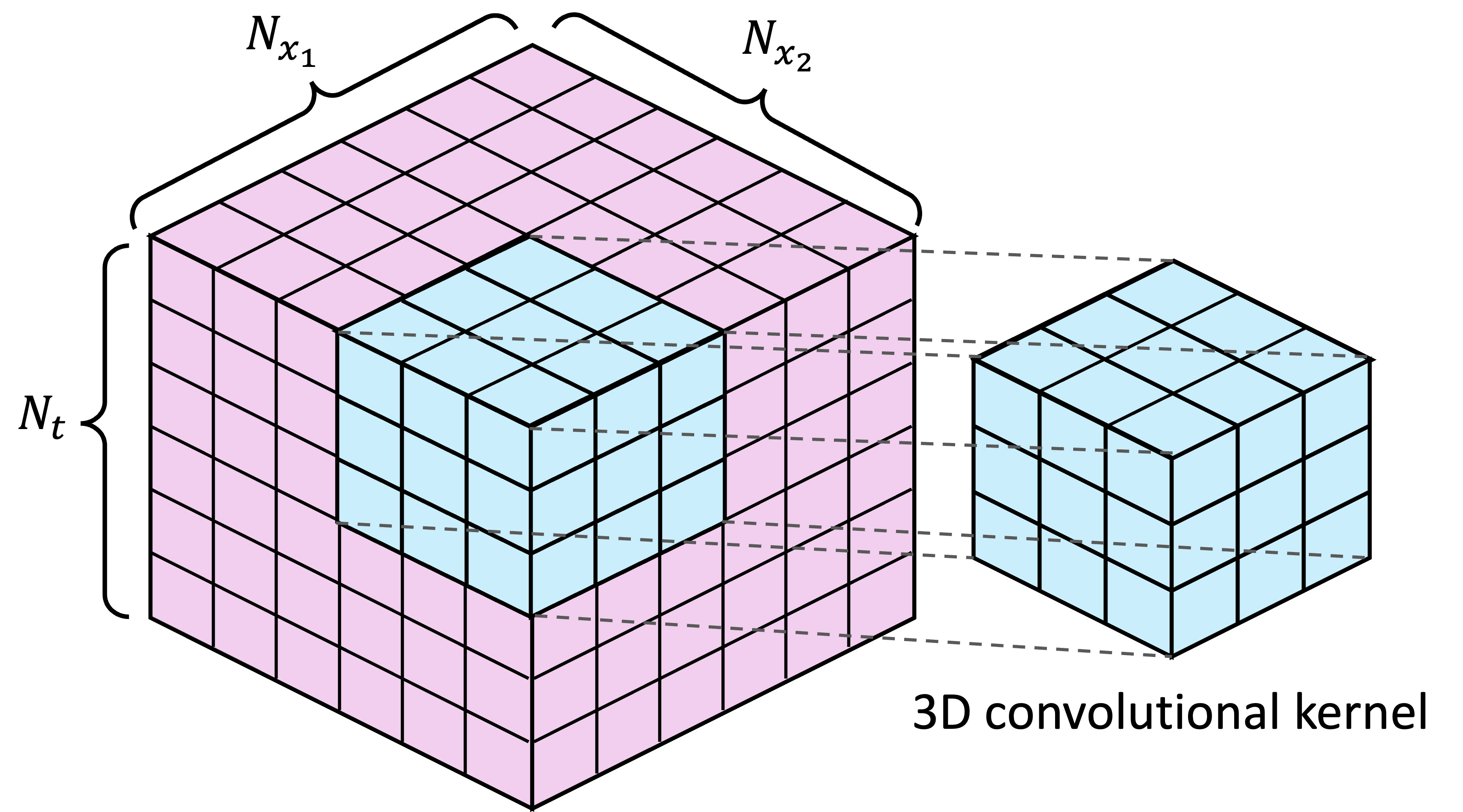}
        \caption{Approach 1}
        \label{fig:conv_approach1}
    \end{subfigure}
    \hfill
    \begin{subfigure}[b]{0.45\textwidth}
        \includegraphics[width=\textwidth]{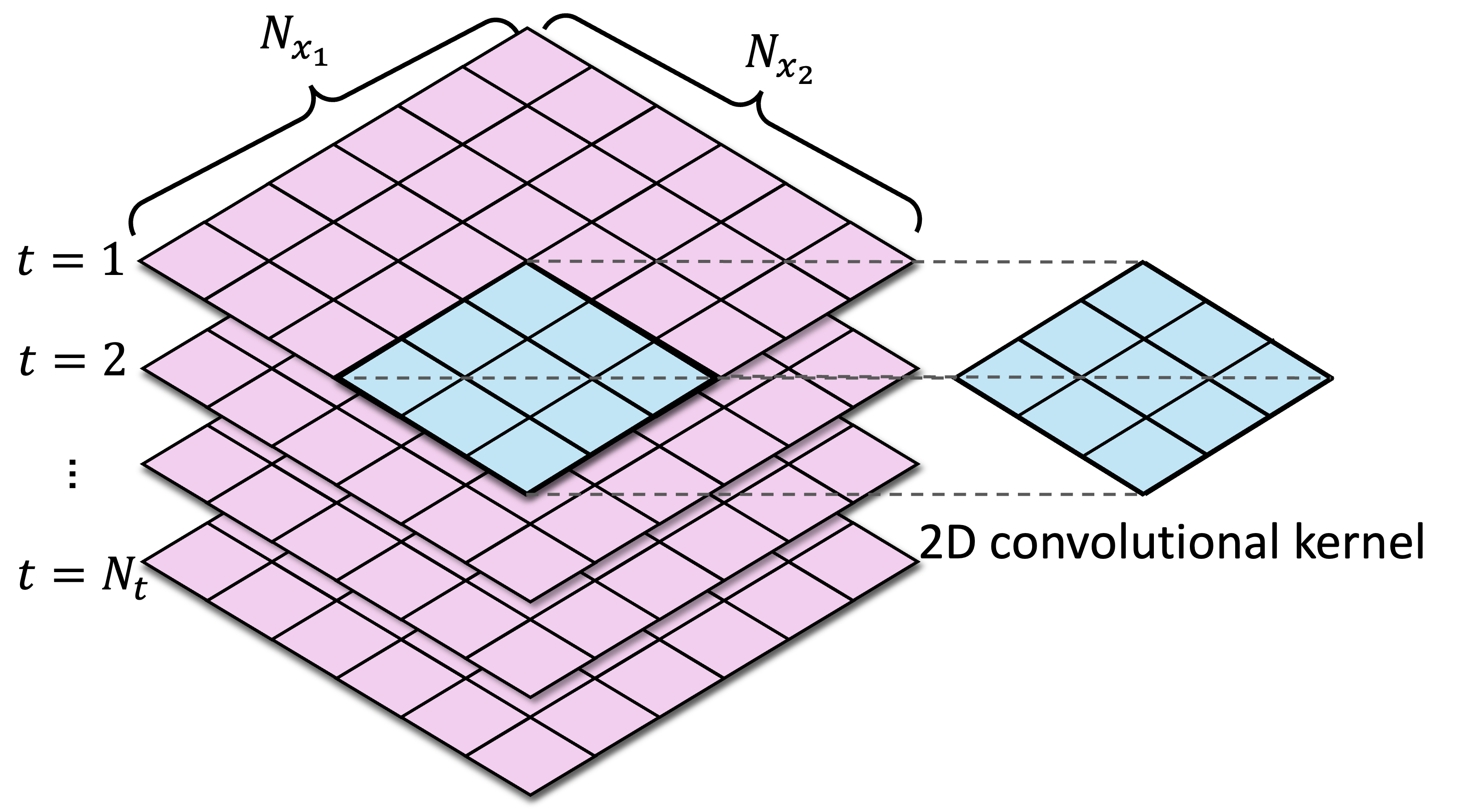}
        \caption{Approach 2}
        \label{fig:conv_approach2}
    \end{subfigure}
    \caption{Illustration of two convolution strategies for modeling space-time-dependent fields: (a) three-dimensional convolution is applied over the spatial and temporal dimensions jointly, and (b) two-dimensional convolutions are applied to each time snapshot.}
    \label{fig:conv_approach}
\end{figure}

Discretization of the space-time-dependent state variable $h$ yields the three-dimensional tensor $\mathbf{h} \in \mathbb{R}^{N_t \times N_{x_1} \times N_{x_2}}$. We consider two approaches for modeling $\mathbf{h}$ with a VAE using three- or two-dimensional convolutional layers. 
In the first approach, we use three-dimensional convolutional layers \cite{ji20123d, tran2015learning, guo2019deep} with the  convolutional kernel $\mathbf{K} \in \mathbb{R}^{k_{x_1} \times k_{x_2} \times k_t}$ as shown in Figure \ref{fig:conv_approach1}. The three-dimensional convolution operation at location $(i, j, k)$ of a tensor $\mathbf{u}$ is given by:
\begin{align}
    (\mathbf{u} * \mathbf{K})[i,j,k] = \sum_{m=0}^{k_{x_1} - 1} \sum_{n=0}^{k_{x_2} - 1} \sum_{p=0}^{k_t - 1} \mathbf{u}[i+m, j+n, k+p] \cdot \mathbf{K}[m,n,p].
\end{align}
In the second approach, we apply two-dimensional convolutions (in the two-dimensional physical space) with the $\mathbf{K} \in \mathbb{R}^{N_{x_1} \times N_{x_2}}$ kernel and model the temporal dimension using input channels \cite{simonyan2014two}. In this approach, the input $\mathbf{h}$ is treated as $N_t$ two-dimensional tensors. The convolution at an output location $(i, j)$ is defined as:
\begin{align}
    (\mathbf{u} * \mathbf{K})[i,j] = \sum_{c=0}^{N_t - 1} \sum_{m=0}^{k_y - 1} \sum_{n=0}^{k_x - 1} \mathbf{u}[c, i+m, j+n] \cdot \mathbf{K}[m,n]
\end{align}
where $c$ is the time step index. This approach performs spatial convolution while adding results across the time dimension.

As noted in \cite{tran2015learning, guo2019deep}, the use of three-dimensional convolutions can better preserve temporal correlations in spatiotemporal data. However, training three-dimensional convolutional neural networks is generally more computationally demanding and susceptible to overfitting, especially when the training dataset is limited in size. Our analysis reveals that the second approach, which utilizes two-dimensional convolutions, offers improved accuracy in reconstructing \( h \). Additionally, it achieves this with a training time that is four times shorter than using three-dimensional convolutions. Therefore, in the numerical examples of this work, we use an $h$-VAE based on the second approach. 

Other design choices in constructing VAEs include the activation function, the number of convolutional layers, the latent space dimensionality, and the type of pooling. 
We adopt the hyperbolic tangent as the activation function. The convolutional channels follow a cascaded increasing/decreasing scheme, where the number of convolutional filters doubles for each successive encoder layer and is halved in the decoder. For the latent space dimensionality, we set $N^y_{\text{latent}} = 150$ for the parameter field $\mathbf{y}$ and $N^h_{\text{latent}} = 90$ for the state field $\mathbf{h}$. These choices are consistent with the number of KLE terms used in the KL-DNN method for the same problem \cite{wang2024bayesian}, where a 150-term truncated KLE retains approximately 93.1\% of the total energy of the $\mathbf{y}$ covariance spectrum, and a 90-term KLE retains approximately 99.955\% of the total energy of the $\mathbf{h}$ covariance spectrum. Regarding the pooling method, for the $y$ field, average pooling yields a lower error than maximum pooling; however, the type of pooling does not significantly impact the reconstruction of the $h$ field. We summarize the $y$-VAE and $h$-VAE hyperparameters and architectures in Table \ref{tab:vae_architecture}.

\begin{table}[!htb]
    \centering
    \caption{Hyperparameters in the  $y$-VAE and $h$-VAE in the Freyberg problem. The input and output to the $y$-VAE are the $1 \times 40 \times 20$ data arrays. The input and output of the $h$-VAE are the $24 \times 40 \times 20$ arrays consisting of the $24$ channels (timesteps).}
    \begin{adjustbox}{max width=\textwidth, max height=\textheight}
    \begin{tabular}{c|c|c|c|c|c|c|c}
        \hline
        & & \textbf{Layer} & \textbf{Kernel Shape} & \textbf{Num of Filters} & \textbf{Input shape} & \textbf{Output shape} & \textbf{Activation} \\
        \hline
        \multirow{13}{*}{\textbf{h-VAE}} 
            & \multirow{8}{*}{Encoder} 
                & Conv2D & (3, 3) & 24 & (24, 40, 20) & (24, 40, 20) & Tanh \\
            & & MaxPool2D & (2, 2) & $\cdots$  & (24, 40, 20) & (24, 20, 10) & $\cdots$ \\
            & & Conv2D  & (3, 3) & 48 & (24, 20, 10) & (48, 20, 10) & Tanh \\
            & & MaxPool2D & (2, 2) & $\cdots$ & (48, 20, 10) & (48, 10, 5) & $\cdots$ \\
            & & Conv2D  & (3, 3)  & 96 & (48, 10, 5) & (96, 10, 5) & Tanh \\
            & & Reshape & $\cdots$  & $\cdots$ & (96, 10, 5) & (4800,) & $\cdots$ \\
            & & Dense (mean) & $\cdots$ & $\cdots$ & (4800,) & $(N_{\text{latent}})$ & Linear \\
            & & Dense (log var) & $\cdots$ & $\cdots$ & (4800,) & $(N_{\text{latent}})$ & Linear \\
        \cline{2-8}
            & \multirow{5}{*}{Decoder} 
                & Dense       & $\cdots$  & $\cdots$  & $(N_{\text{latent}})$  & (4800,)  &  Linear \\ 
            & & Reshape     & $\cdots$  & $\cdots$  & (4800,) & (96, 10, 5)  & $\cdots$ \\
            & & ConvTranspose2D & (4, 4) & 48 & (96, 10, 5)  & (48, 20, 10)  & Tanh \\ 
            & & ConvTranspose2D & (4, 4) & 24 & (48, 20, 10) & (24, 40, 20) & Tanh \\ 
            & & Conv2D      & (3, 3) & 24  & (24, 40, 20)  & (24, 40, 20) & Linear \\
        \hline
        \hline
            \multirow{13}{*}{\textbf{y-VAE}} 
            & \multirow{8}{*}{Encoder} 
                & Conv2D & (3, 3) & 8  & (1, 40, 20) & (8, 40, 20) & Tanh \\
            & & AvgPool2D & (2, 2) & $\cdots$  & (8, 40, 20) & (8, 20, 10) & $\cdots$ \\
            & & Conv2D  & (3, 3) & 16 & (8, 20, 10) & (16, 20, 10) & Tanh \\
            & & AvgPool2D & (2, 2) & $\cdots$ & (16, 20, 10) & (16, 10, 5) & $\cdots$ \\
            & & Conv2D  & (3, 3) & 32 & (16, 10, 5) & (32, 10, 5) & Tanh \\
            & & Reshape & $\cdots$  & $\cdots$ & (32, 10, 5) & (1600,) & $\cdots$ \\
            & & Dense (mean) & $\cdots$ & $\cdots$ & (1600,) & $(N_{\text{latent}})$ & Linear \\
            & & Dense (log var) & $\cdots$ & $\cdots$ & (1600,) & $(N_{\text{latent}})$ & Linear \\
        \cline{2-8}
            & \multirow{5}{*}{Decoder} 
                & Dense & $\cdots$ & $\cdots$ & $(N_{\text{latent}})$ & (1600,) & Linear \\
            & & Reshape  & $\cdots$  & $\cdots$  & (1600,)   & (32, 10, 5)  & $\cdots$ \\
            & & ConvTranspose2D & (3, 3) & 16 & (32, 10, 5) & (16, 10, 5) & Tanh  \\
            & & ConvTranspose2D & (3, 3) & 8 & (16, 10, 5) & (8, 20, 10) & Tanh  \\
            & & ConvTranspose2D & (3, 3) & 1 & (8, 20, 10) & (1, 40, 20) & Linear  \\
        \hline
    \end{tabular}
    \end{adjustbox}
    \label{tab:vae_architecture}
\end{table}

\section{Implementation details of the Fourier Neural Operator}\label{sec:appendix_fno}

We considered two approaches for applying FNO to the time-dependent Freybers problem: (1) treating time as an additional physical dimension, or (2) treating time as channels. We observed that the second approach yields a two times smaller $rl^2_h$ error and requires 12 times less training time than the first approach. Therefore, we restrict our further discussion to the second configuration.
Each parameter sample $\mathbf{y} \in \mathbb{R}^{ N_{x_1} \times N_{x_2}}$ is augmented to a tensor $\tilde{\mathbf{y}} \in \mathbb{R}^{ 3 \times N_{x_1} \times N_{x_2}}$ as input, where the three channels contain the $y$ value at the $(i,j)$ element and its $x_1$ and $x_2$ coordinates. Then, $\tilde{\mathbf{y}}$ is passed through a shallow DNN $P: \tilde{\mathbf{y}} \rightarrow \mathbf{v}_0$, where $\mathbf{v}_0 \in \mathbb{R}^{N_v \times N_{x_1} \times N_{x_2}}$. This lifts the channel dimension of $\tilde{\mathbf{y}}$ from 3 to $N_v>>3$. Next, a sequence of $N$ Fourier layers is applied: $\mathbf{v}_0 \rightarrow \mathbf{v}_1 \rightarrow ... \rightarrow \mathbf{v}_N$. 

\begin{figure}[!htb]
\centering
\includegraphics[width=0.85\linewidth,keepaspectratio]{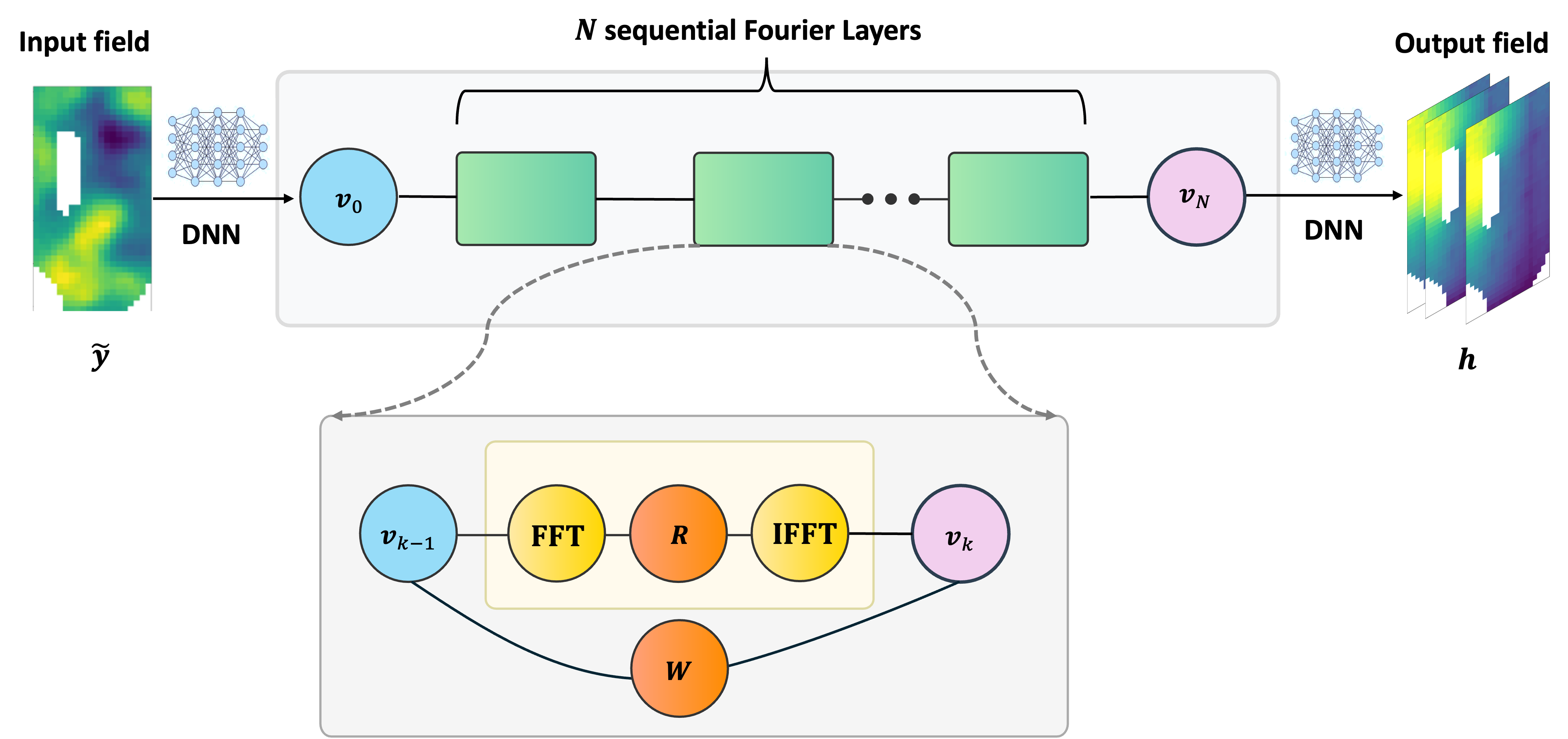}
\caption{ FNO schematics for the Freyberg problem (modified after~\cite{li2020fourier}): the input $\tilde{\mathbf{y}}$ is lifted to $\mathbf{v}_0$-dimensional space using a fully-connected DNN. $N$ sequential Fourier layers are applied, in each layer (the $k$-th layer is shown), the FFT is applied to the input. A linear transformation $\mathbf{R}$ is applied to the lower Fourier modes, filtering out high-frequency components. The IFFT transforms the data back to the physical domain. Additionally, a local linear transformation is applied directly to the input of each Fourier layer, and its output is added to the result of the IFFT, ensuring that both global and local features are captured. Finally, the output from the last Fourier layer, $\mathbf{v}_N$, is projected back to the target output space using another fully connected DNN.}
\label{fig:fno_structure}
\end{figure}

Within each Fourier layer, we first apply the fast Fourier transform (FFT) $F$ to the data, followed by a linear transform $R$ on the lower Fourier modes. Then, we filter out the higher modes and apply the inverse fast Fourier transform (IFFT) $F^{-1}$. Next, we apply a local linear transform $W$ to the input added to the output of the IFFT. For the $k$-th Fourier layer, the output is represented as:
\begin{eqnarray}\label{eq:FNO}
    \mathbf{v}_k = \sigma \left [ F^{-1}(R_k(F(\mathbf{v}_{k-1}))) + W_k\mathbf{v}_{k-1} \right]
\end{eqnarray}
Finally,  we process $\mathbf{v}_N$ with another fully connected neural network $Q: \mathbf{v}_N \rightarrow h$, which projects the channel dimension of $\mathbf{v}_N$ back to the number of timesteps. Figure \ref{fig:fno_structure} shows the schematics of FNO for the Freyberg problem.

In our implementation, we use four sequential Fourier layers with a feature dimension of $N_v = 128$. We find that further increasing $N_v$ significantly lengthens training time without reducing testing error. We filter out Fourier modes with wavenumbers greater than eight. The hyperbolic tangent is employed as the activation function throughout the network.

\section{Implementation details of the DeepONet}\label{sec:appendix_deeponet}

DeepONet consists of the branch and trunk networks \cite{lu2021learning}. The branch network encodes the input function, while the trunk network captures spatial (and, if applicable, temporal) information by learning a set of basis functions for the target space. The outputs of the two subnetworks are combined via a dot product to approximate the target operator as
\begin{eqnarray}\label{eq:vanilla_deeponet}
    h(\mathbf{x}, t) = \mathcal{G}(\mathbf{y})(\mathbf{x}, t) \approx \sum_{k = 1}^{p}\mathcal{NN}^{(k)}_{\text{branch}}(\mathbf{y}; \theta_{\text{branch}})\cdot \mathcal{NN}^{(k)}_{\text{trunk}}(\mathbf{x}, t; \theta_{\text{trunk}}) + b_0
\end{eqnarray}
where $\mathcal{NN}^{(k)}_{\text{branch}}$ and $\mathcal{NN}^{(k)}_{\text{trunk}}$ denote the $k$-th components of the outputs of the branch and trunk networks, respectively. 

In the standard DeepONet formulation, the discretized parameter field is used as input to the branch network. For the Freyberg problem, the input $y$ field is discretized in the two-dimensional array $\mathbf{y}$, as illustrated in Figure~\ref{fig:freyberg_conceptual}. The $\mathbf{y}$ array can be flattened into a one-dimensional vector $[y_1, \dots, y_{N_e}]^T$ ($N_e$ is the size of the $\bm{y}$ array) and used as an input to a fully connected branch network. Alternatively, the branch can be modeled using a CNN with $\mathbf{y}$ as the input. In this work, we employ a truncated KLE 
\begin{align}
  y(\mathbf{x}) \approx \overline{y}(\mathbf{x}) + \sum_{i=1}^{N_\xi} \sqrt{\lambda_{i} }\phi_i(\mathbf{x}) \xi_i
\end{align}
to reduce the dimensionality of $\mathbf{y}$, 
where $\overline{y}$ is the mean function, $\lambda_i$ and $ \phi_i$ are the eigenvalues and eigenfunctions of the $y$ covariance computed from training data ($\lambda_i$ are arranged in the descending order), $N_\xi$ is the number of terms in the truncated KLE, and $\xi_i$ are the KLE coefficients.
We use the vector of KLE coefficients $\boldsymbol{\xi}$ in place of the whole field $y(\mathbf{x})$ as input to the branch DNN network. 
 The mean function and the covariance are computed from the training data:
\begin{equation}\label{eq:ens_mean}
  \overline{y}(\mathbf{x})\approx \frac{1}{N_{\text{train}}} \sum_{i=1}^{N_{\text{train}}} y^{(i)}(\mathbf{x}) , 
\end{equation}
and
\begin{equation}\label{eq:ens_cov}
  C_y(\mathbf{x}, \mathbf{x}') \approx \frac{1}{N_{\text{train}}-1} \sum_{i=1}^{N_{\text{train}}} \left \{ \left[y^{(i)} ( \mathbf{x})-\overline{y}(\mathbf{x})\right] \left[ y^{(i)}(\mathbf{x}') -\overline{y}(\mathbf{x}')\right] \right \}.
\end{equation}
Then, $\phi_y^{i}(\mathbf{x})$ and $\lambda_y^{i}$ are found by solving the eigenvalue problem
\begin{align}\label{eq:kle_y_eigenvalue}
\int_\Omega C_y(\mathbf{x}, \mathbf{x}') \phi_y^{i}(\mathbf{x}') \, d\mathbf{x}' = \lambda_y^{i} \phi_y^{i}(\mathbf{x}), \quad i = 1, \dots, N_e.
\end{align}

\begin{figure}[!htb]
\centering
\includegraphics[width=0.65\linewidth,keepaspectratio]{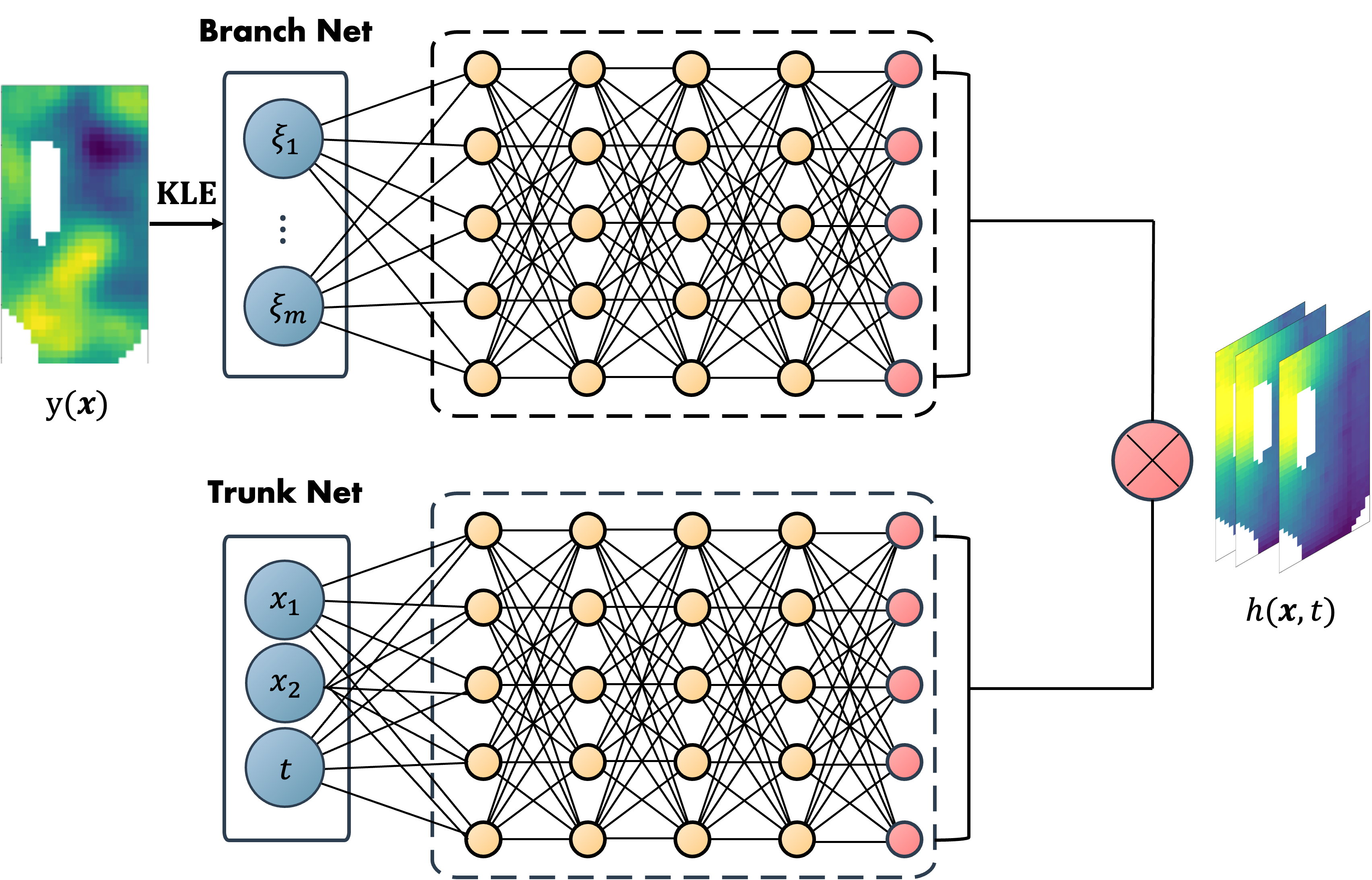}
\caption{ DeepONet model for the Freyberg problem. The coefficients in the KLE of $y$ are used as the input to the branch network. The inputs to the trunk DNN are the space and time coordinates.}
\label{fig:deeponet_structure}
\end{figure}

The number of KLE terms $N_{\xi}$ is selected according to the condition:
\begin{align}
\sum_{i=N_{\xi}+1}^{N_e} \lambda^i_y \leq \text{rtol} \sum_{i=1}^{N_e} \lambda^i_y,
\end{align}
where rtol is the prescribed tolerance. In this 
In this work, we set $\text{rtol} = 0.07$, resulting in $N_\xi = 150$. 
We use four hidden layers, each with 1200 neurons, in the branch DNN. The trunk DNN consists of four hidden layers, each with 300 neurons. 
For a fair comparison of the DeepONet and VAE-DNN models, the size of the output layers of the trunk and branch networks is set to 90, corresponding to the number of latent variables in the $h$-VAE of the VAE-DNN model. 
Figure \ref{fig:deeponet_structure} shows the schematic representation of the DeepONet model for the Freyberg problem. 

\end{document}